\documentclass{article}

\usepackage{microtype}
\usepackage{graphicx}
\usepackage{subcaption}
\usepackage{booktabs} 

\usepackage{hyperref}

\usepackage[preprint]{icml2026}

\usepackage{amsmath}
\usepackage{amssymb}
\usepackage{mathtools}
\usepackage{amsthm}
\usepackage{booktabs}
\usepackage{tabularx}
\usepackage{multirow}
\usepackage{array}
\usepackage{tcolorbox}

\newcolumntype{C}{>{\centering\arraybackslash}X}

\usepackage[capitalize,noabbrev]{cleveref}

\theoremstyle{plain}

\theoremstyle{definition}

\theoremstyle{remark}

\usepackage[textsize=tiny]{todonotes}

\icmltitlerunning{Follow the Clues, Frame the Truth: Hybrid-evidential Deductive Reasoning in OV-MER}

\begin{document}

\twocolumn[
  \icmltitle{Follow the Clues, Frame the Truth: Hybrid-evidential Deductive Reasoning in Open-Vocabulary Multimodal Emotion Recognition}



  \icmlsetsymbol{equal}{*}

  \begin{icmlauthorlist}
    \icmlauthor{Yu Liu}{equal,yyy}
    \icmlauthor{Lei Zhang}{equal,yyy}
    \icmlauthor{Haoxun Li}{yyy}
    \icmlauthor{Hanlei Shi}{yyy}
    \icmlauthor{Yuxuan Ding}{yyy}
    \icmlauthor{Leyuan Qu}{yyy}
    \icmlauthor{Taihao Li}{yyy}
  \end{icmlauthorlist}

  \icmlaffiliation{yyy}{Hangzhou Institute for Advanced Study, University of Chinese Academy of Sciences, Hangzhou, China}

  \icmlcorrespondingauthor{Taihao Li}{lith@ucas.ac.cn}
  \icmlcorrespondingauthor{Leyuan Qu}{leyuan.qu@ucas.ac.cn}

  \icmlkeywords{Machine Learning, ICML}

  \vskip 0.3in
]



\printAffiliationsAndNotice{}  

\begin{abstract}
Open-Vocabulary Multimodal Emotion Recognition (OV-MER) is inherently challenging due to the ambiguity of equivocal multimodal cues, which often stem from distinct unobserved situational dynamics. 
While Multimodal Large Language Models (MLLMs) offer extensive semantic coverage, their performance is often bottlenecked by premature commitment to dominant data priors, resulting in suboptimal heuristics that overlook crucial, complementary affective cues across modalities.
We argue that effective affective reasoning requires more than surface-level association; it necessitates reconstructing nuanced emotional states by synthesizing multiple evidence-grounded rationales that reconcile these observations from diverse latent perspectives.
We introduce \textbf{HyDRA}, a \textbf{Hy}brid-evidential \textbf{D}eductive \textbf{R}easoning \textbf{A}rchitecture that formalizes inference as a Propose–Verify–Decide protocol. 
To internalize this abductive process, we employ reinforcement learning with hierarchical reward shaping, aligning the reasoning trajectories with final task performance to ensure they best reconcile the observed multimodal cues. 
Systematic evaluations validate our design choices, with HyDRA consistently outperforming strong baselines—especially in ambiguous or conflicting scenarios—while providing interpretable, diagnostic evidence traces.
\end{abstract}

\section{Introduction}
A girl stands on a podium holding a silver medal, her eyes filled with tears. A standard multimodal classifier may quickly commit to \textit{a single dominant label} such as \textit{sadness}. Yet, these equivocal multimodal cues can support a plurality of co-occurring interpretations: for instance, \textit{sadness} intertwined with \textit{pride} (achievement), \textit{regret} (missing the gold), or \textit{relief} (concluding a struggle). This is not an edge case; inferring affective states is inherently challenging because emotion is often contextually underspecified by surface signals, requiring the inference of unobserved situational dynamics~\cite{poria2017review, lian2023explainable}.

Conventional Multimodal Emotion Recognition (MER) models have been restricted to fixed label taxonomies, which struggle to reflect the open-ended and nuanced nature of human affect~\cite{han2025benchmarking, zhao2025multimodal}. 
Open-Vocabulary MER (OV-MER) relaxes this constraint and enables flexible predictions beyond fixed taxonomies~\cite{lian_ovmer, lian2023explainable}. 
However, OV-MER introduces a key mismatch: evaluation stresses label cardinality and synonymy, while training often reduces to token-level likelihoods that ignore such semantic structure~\cite{bhattacharyya-wang-2025-evaluating}. 
As a result, models exploit shortcut heuristics and become fragile under cross-modal conflict~\cite{geirhos2020shortcut, li-etal-2023-evaluating, zhu2025mitigating}.
This gap naturally motivates adopting stronger semantic priors to interpret open-ended labels and latent contexts.

To supply such priors, recent work increasingly adopts Multimodal Large Language Models (MLLMs) as general-purpose multimodal reasoners~\cite{liu2023visual, lin2024video, ge2024video}. 
They offer broad semantic coverage, yet they often fail in the same way: \emph{premature commitment} to a dominant, prior-driven interpretation, with complementary cues left unused~\cite{bai2024hallucination, leng2024mitigating, lei2024large}.
Standard grounding strategies therefore suppress generation to mitigate hallucinations~\cite{dhuliawala2024chain}. 
We argue this misses the point: the main issue is not generation itself, but \emph{biased singular interpretation from incomplete evidence}~\cite{zhang2025simignore}.
Under ambiguous or conflicting cues, models collapse to a single dominant narrative and ignore complementary affective evidence across modalities—a phenomenon mirroring the ``System 1" thinking that lacks the rigors of deliberative verification~\cite{kahneman2011thinking}.

Our key idea is to treat the generative prior as structured intuition, not as a final decision.
Instead of accepting a single rationale, we generate multiple competing candidates and then \emph{force} them to be adjudicated by the observed multimodal evidence~\cite{mistretta2025cross, wu2025antidote, chang2026abductivemllmboostingvisualabductive}.
We introduce \textbf{HyDRA}, a \textbf{Hy}brid-evidential \textbf{D}eductive \textbf{R}easoning \textbf{A}rchitecture that formalizes this process as a \textbf{Propose--Verify--Decide} protocol.
HyDRA \textbf{proposes} a small set of diverse situational hypotheses to avoid early collapse to a single, prior-driven narrative. 
It then \textbf{verifies} them through evidence-constrained comparison, eliminating candidates that conflict with salient multimodal observations. 
Finally, it \textbf{decides} by selecting the hypothesis that best reconciles the observed cues, effectively retaining evidence-consistent affective descriptors while filtering shallow semantics and spurious priors.

To make this behavior a learned capability rather than a prompting trick, we optimize HyDRA with reinforcement learning and hierarchical reward shaping.
Using Group Relative Policy Optimization (GRPO)~\cite{shao2024deepseekmathpushinglimitsmathematical, guo2025deepseek}, we align the policy to favor diverse proposal generation, rigorous evidence grounding, and accurate task-specific decisions.
GRPO acts as a differential filter that rewards evidential closure, pushing the model away from plausible-sounding hallucinations and toward rationales necessitated by the multimodal evidence~\cite{li2025videohallu}.
This also yields diagnostic reasoning traces for analyzing model behavior under ambiguity and conflict.

\textbf{Contributions:} 
\begin{itemize}
    \item \textbf{A hypothesis-driven inference interface for OV-MER.} We formalize OV-MER as a Propose–Verify–Decide procedure that generates multiple latent-context hypotheses and performs evidence-constrained adjudication to avoid premature commitment under equivocal multimodal cues.
    \item \textbf{Learning to adjudicate, not prompting to look structured.} We couple the protocol with GRPO-based policy optimization and hierarchical rewards to internalize comparative verification and evidence closure, outperforming prompt-only and alternative training paradigms under the same backbone.
    \item \textbf{Systematic evidence beyond aggregate scores.} We provide controlled ablations on hypothesis cardinality, reward components, and training paradigms, and demonstrate that the gains are driven by multi-path adjudication rather than model scale.
\end{itemize}

\section{Related Work}

\subsection{From Closed-set to Open-Vocabulary MER}
Conventional MER typically treats affect as a classification task within fixed taxonomies~\cite{zadeh2016mosimultimodalcorpussentiment}. While structured, these paradigms struggle to capture the fluid, overlapping, and context-dependent nature of human emotions in the wild~\cite{guerdelli2023towards}. To address this, OV-MER has emerged, leveraging unrestricted natural language to bridge discrete labels with continuous semantic spaces~\cite{lian_ovmer}.

Despite this flexibility, OV-MER faces a persistent challenge in resolving situational dynamics under modal ambiguity or conflict~\cite{10.1145/3746027.3754856}. When visual and acoustic signals contradict—such as a ``tearful smile"—current models often lack the inferential depth to deduce the latent context, resulting in superficial mappings between signals and descriptors~\cite{wei-etal-2023-tackling}. While MLLMs like LLaVA-NeXT and InternVL3.5 offer vast world knowledge for affective interpretation~\cite{li2024llavanextinterleavetacklingmultiimagevideo, wang2025internvl3_5}, their performance is frequently bottlenecked by a ``premature commitment" to dominant linguistic priors~\cite{zhou2023analyzing, huang2024opera, goyal2017making}. This leads to suboptimal heuristics that ignore subtle, complementary cues, necessitating a more rigorous, evidence-grounded reasoning framework~\cite{dhuliawala2024chain, zelikman2024quiet}.

\subsection{Heuristic Bias and Premature Commitment in MLLMs}
The ``premature commitment" observed in MLLMs is often attributed to a reliance on strong internal priors over sparse external evidence~\cite{qian2025prismbenchbenchmarkpuzzlebasedvisual,poria2017review, lian2023explainable}, a phenomenon widely documented in complex multimodal reasoning. In affective contexts, this manifests as a collapse into a single dominant narrative, disregarding the ``long-tail" of nuanced emotional possibilities~\cite{rha2025learningattendfirstmodalityimportanceguided}. Recent studies suggest that standard grounding techniques often fail to mitigate this bias because they focus on token-level alignment rather than higher-order semantic consistency~\cite{wu2025grounded, mohsin2025fundamental}.

We argue that robust affective reasoning requires an abductive process that transcends surface-level associations~\cite{chang2026abductivemllmboostingvisualabductive}. This perspective is motivated by recent findings that large language models often struggle with logical parsimony in abductive reasoning, frequently failing to reconcile evidence through the simplest or most consistent explanatory paths~\cite{sun2025languagemodelsfollowoccams,zhang2025simignore}. Specifically, synthesizing multiple evidence-grounded rationales—rather than pursuing a singular path—has been shown to reconcile conflicting observations across diverse latent perspectives in other reasoning-heavy tasks like video understanding and fact-checking~\cite{dang2025mupamultipathagenticreasoning,zhao2023verify}. By formalizing this as a Propose–Verify–Decide protocol, HyDRA ensures that the final interpretation is not merely a reflection of model priors, but a conclusion necessitated by the totality of multimodal evidence.

\begin{figure*}[ht]
  \vskip 0.2in
  \begin{center}
    \centerline{\includegraphics[width=0.92\textwidth]{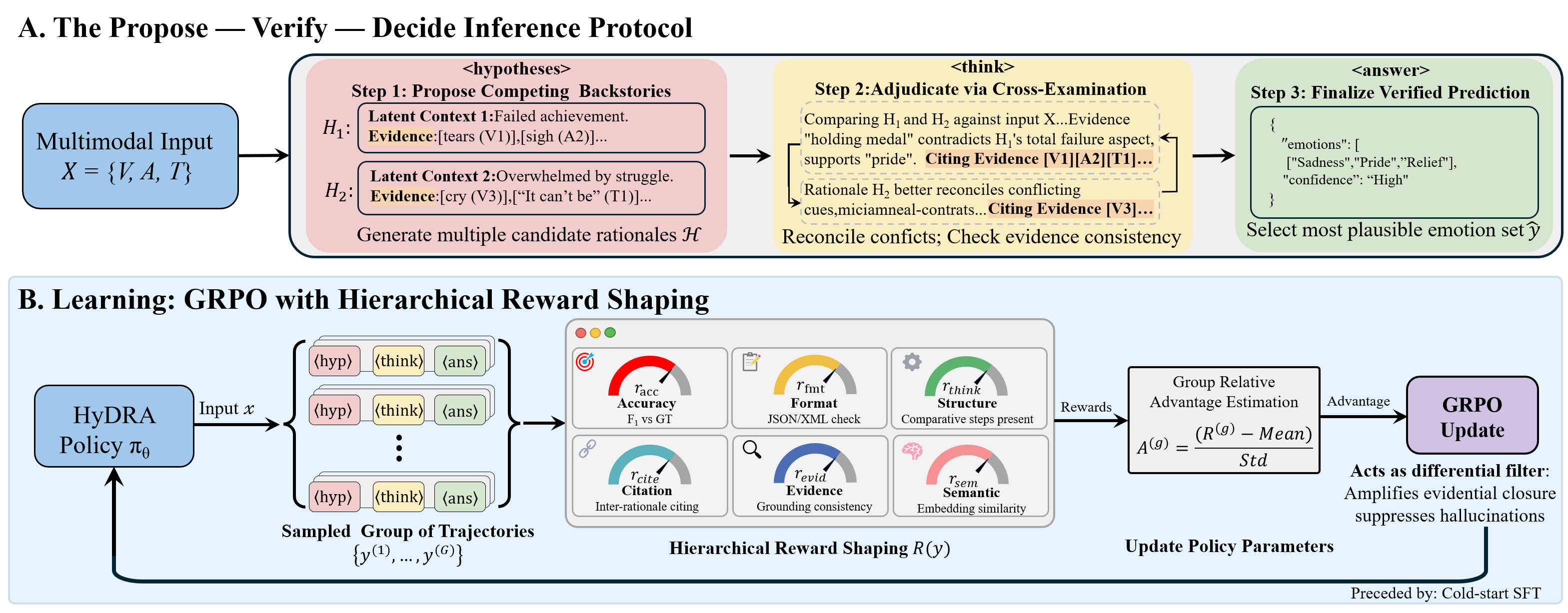}}
    \caption{
      \textbf{HyDRA overview.} 
(\textbf{A}) \textit{Propose--Verify--Decide} inference protocol: given multimodal input, HyDRA proposes multiple latent-context hypotheses, adjudicates them via evidence-constrained comparison with explicit citations, and outputs the most plausible emotion set. 
(\textbf{B}) Learning HyDRA with GRPO and hierarchical reward shaping: for each input we sample a group of structured trajectories in the \texttt{<hyp>--<think>--<ans>} format, score them with six rewards, compute group-relative advantages, and update the policy to favor \textit{evidential closure} and robust decisions under ambiguity. Textual rationales are grounded to human-verified textualized multimodal cues (ObsG) provided by the datasets, and enforced via semantic alignment to ground-truth cue descriptions.
    }
    \label{fig:overview}
  \end{center}
\end{figure*}

\subsection{Evidence-Grounded Reasoning and Rationale Synthesis}
We argue that effective affective reasoning necessitates moving beyond surface-level associations, as emotions are often contextually underspecified by surface signals~\cite{poria2017review, lian2023explainable}. While MLLMs can generate descriptive labels, they frequently fall victim to shortcut learning, where the model bypasses deep situational analysis by mapping salient but equivocal features (e.g., ``tears") directly to high-frequency labels (e.g., ``sadness") based on statistical co-occurrences in training data~\cite{sun-etal-2024-exploring,geirhos2020shortcut, li-etal-2023-evaluating, zhu2025mitigating}. Such associations are often ``hijacked" by dominant linguistic priors, leading to systematic failures in ``emotion conflict" scenarios where surface signals contradict latent situational dynamics~\cite{wu2025spurioussignalsdebiasingmultimodal}. Consequently, simple associative mapping is insufficient for capturing evoked emotions—those deeply rooted in situational context rather than mere facial muscle movements~\cite{doi:10.1177/1529100619832930}.

To overcome these heuristic bottlenecks, robust reasoning requires reconstructing nuanced emotional states by synthesizing multiple evidence-grounded rationales that reconcile observations from diverse latent perspectives~\cite{wang2022self, zhao2023verify, dhuliawala2024chain}. This perspective is rooted in abductive reasoning, which requires the generation of ``Explanatory Hypotheses" to account for incomplete or ambiguous visual evidence~\cite{chang2026abductivemllmboostingvisualabductive}. 
Formulated as Propose–Verify–Decide, HyDRA makes affective prediction the outcome of comparative verification and elimination under multimodal constraints, yielding decisions that remain stable under ambiguity and cross-modal conflict.

\section{Methodology}
\label{sec:methodology}

\subsection{Overview}
We introduce HyDRA, a framework that operationalizes multimodal emotion reasoning as a structured Propose–Verify–Decide sequence. As shown in Fig.~\ref{fig:overview}, HyDRA replaces the standard singular discriminative path with a multi-path reconstruction process, synthesizing rationales from divergent perspectives to mitigate prior-driven biases. The framework is optimized via a two-stage pipeline: (i) \textbf{Cold-start Multimodal Supervision} to seed the structured reasoning protocol, and (ii) \textbf{Policy Optimization via GRPO} to enforce multimodal grounding and deductive consistency through hierarchical reward shaping.

\subsection{The HyDRA Protocol}
\label{sec:protocol}
Unlike standard MLLMs, HyDRA treats emotion recognition as a reconstruction task over a latent situational space $\mathcal{S}$. Given multimodal observations $\mathcal{X}=\{V, A, T\}$, we estimate the optimal emotion set $\hat{\mathcal{Y}}$ by adjudicating over $K$ generated rationales $\mathcal{H} = \{H_1, \dots, H_K\}$:
\begin{equation}
\label{eqa:overall_presentation}
\hat{\mathcal{Y}} = \text{Decide} \left( \sum_{k=1}^K \Phi(H_k, \mathcal{X}) \cdot \Psi(H_k, \mathcal{Y}) \right),
\end{equation}
where $\Phi(\cdot)$ measures the abductive grounding of rationale $H_k$ against $\mathcal{X}$, and $\Psi(\cdot)$ represents the deductive consistency between the rationale and label $\mathcal{Y}$. This process is operationalized through a three-stage inference interface:

\begin{itemize}
    \item \textbf{Proposal (Abductive Stage)}: The model generates $K$ competing hypotheses $\mathcal{H}$, where each $H_k$ encapsulates a latent context $s_k$ and predicted cue descriptions $E_k = \{e_{k,1}, e_{k,2}, \dots\}$.
    \item \textbf{Verification (Deductive Stage)}: The model performs a ``cross-examination" within a $\langle \texttt{think} \rangle$ block, treating each $H_k$ as a premise to verify its consistency with observations $\mathcal{X}$.
    \item \textbf{Decision}: In HyDRA, the $\text{Decide}$ operation is operationalized as the final synthesis in the $\langle \texttt{think} \rangle$ block, where the model identifies the optimal rationale $H^*$ that maximizes the joint grounding strength $\Phi$ and consistency $\Psi$, thereby yielding the final emotion set $\hat{\mathcal{Y}}$.
\end{itemize}

While Eq.~\ref{eqa:overall_presentation} provides a theoretical objective, HyDRA operationalizes it end-to-end: $\Phi$ is implicitly optimized via semantic alignment of predicted cues, and $\Psi$ is enforced through task-level rewards during Reinforcement Learning (RL) stage. 

\subsection{Cold-Start Multimodal Supervision}
To initialize the reasoning protocol $\pi_\theta$, we perform supervised fine-tuning (SFT) on a corpus of structured reasoning traces. The model comprises a causal transformer backbone integrated with vision and audio encoders via projection layers. Given multimodal inputs $\mathcal{X}$, we minimize the negative log-likelihood over the expert trace $\mathcal{Y}$:
\begin{equation}
\mathcal{L}_{\text{SFT}}(\theta) = - \mathbb{E}_{(\mathcal{X}, \mathcal{Y}) \sim \mathcal{D}_{\text{SFT}}} \sum_{t=1}^{|\mathcal{Y}|} \log p_\theta(y_t \mid \mathcal{X}, y_{<t}),
\end{equation}
where the loss is restricted to the assistant’s tokens, effectively "seeding" the Propose–Verify–Decide schema into the policy.

\subsection{Policy Optimization via GRPO}

To internalize the abductive protocol, we employ GRPO~\cite{shao2024deepseekmathpushinglimitsmathematical}. Unlike standard reinforcement learning, GRPO estimates advantage by comparing a group of $G$ trajectories $\{y^{(1)}, \dots, y^{(G)}\}$ sampled from the same prompt $\mathcal{X}$. The advantage $A^{(g)}$ for each candidate is normalized against the group mean:
\begin{equation}
A^{(g)} = \frac{R(y^{(g)}) - \text{mean}({R(y^{(i)})}_{i=1}^G)}{\text{std}({R(y^{(i)})}_{i=1}^G) + \epsilon}.
\end{equation}

\textbf{GRPO as a Differential Filter.} This mechanism provides a critical inductive bias for HyDRA: by comparing $G$ divergent paths in a single step, the advantage $A^{(g)}$ serves as a differential filter that rewards evidential closure. Specifically, trajectories that successfully synthesize conflicting cues (higher $\Phi \cdot \Psi$ in Eq.~\ref{eqa:overall_presentation}) are amplified, while those collapsing into biased priors are suppressed.

The policy is updated by minimizing:
\begin{equation}
\mathcal{L}_{\text{GRPO}}(\theta)
= -\frac{1}{G}\sum_{g=1}^G
\Big(\mathbb{E}_{t}\!\left[r_{g,t}(\theta)\,A^{(g)}\right]
-\beta\,\mathrm{KL}(\pi_\theta\|\pi_{\text{ref}})\Big).
\end{equation}
where the policy‑gradient term $\mathbb{E}_{t}\!\left[r_{g,t}(\theta)\,A^{(g)}\right]$ is a token-wise importance-weighted policy gradient without ratio clipping, $\mathrm{KL}(\pi_\theta\|\pi_{\text{ref}})$ denotes the KL divergence to the reference policy, and
$r_{g,t}(\theta)=\exp\!\Big(\log p_\theta(y^{(g)}_t|\mathcal{X})
-\mathrm{sg}\big(\log p_\theta(y^{(g)}_t|\mathcal{X})\big)\Big)$ uses stop-gradient $\mathrm{sg}(\cdot)$. Through iterative optimization, the model learns to prioritize evidence-grounded synthesis over surface-level association, as the latter consistently yields lower relative rewards within a group.

\subsection{Hierarchical Reward Shaping}
The core of HyDRA's reliability is a hierarchical reward function $R$ that jointly constrains structural adherence, evidence grounding, and task performance (detailed weights and mathematical formulations are provided in Appendices~\ref{app:reward-weights} and~\ref{app:reward_detailed_formulations}):
\begin{equation}
\begin{aligned}
   R = &\lambda_{\text{acc}} r_{\text{acc}} + \lambda_{\text{fmt}} r_{\text{fmt}} + \lambda_{\text{think}} r_{\text{think}} \\ &+ \lambda_{\text{cite}} r_{\text{cite}} + \lambda_{\text{evid}} r_{\text{evid}} + \lambda_{\text{sem}} r_{\text{sem}}. 
\end{aligned}
\end{equation}

\textbf{Evidential closure has two complementary forms.} (1) Intra-trace closure enforces that claims in $\langle \texttt{think} \rangle$ can be traced back to the self-declared evidence pool extracted from $\langle \texttt{hypotheses} \rangle$, preventing the model from introducing unsupported cues ($r_{\text{evid}}$, $r_{\text{cite}}$).
(2) Annotation-grounded closure aligns the predicted cue descriptions with human-verified multimodal cue annotations provided by the datasets ($r_{\text{sem}}$), serving as an external grounding proxy for multimodal faithfulness.
We therefore use “grounded” to denote grounding to verified textualized multimodal cues, rather than pixel-level verification. 

\begin{table*}[t]
\caption{Main results on general and open-vocabulary emotion recognition. We compare HyDRA against several multimodal baselines (up to 7B). $S_1$ and $S_2$ denote coarse- and fine-grained F1-scores on the OV-FG task, respectively. Despite its smaller scale, HyDRA achieves the best average performance and significantly outperforms 7B models on open-vocabulary reasoning. Best results are \textbf{bolded}; second-best are \underline{underlined}. $^*$: AffectGPT trained on OV-MERD~\cite{lian2024ov}.}
\label{tab:metrics_track2}
\begin{center}
\begin{footnotesize}
\begin{sc}
\renewcommand{\arraystretch}{1}
\newcolumntype{C}{>{\centering\arraybackslash}X}
\begin{tabularx}{\linewidth}{l C C C C C C C C}
\toprule
\multirow{2}{*}{Model} & \multirow{2}{*}{\#Params}
& \multicolumn{2}{c}{Basic}
& \multicolumn{2}{c}{Sentiment}
& \multicolumn{2}{c}{OV-FG}
& \multirow{2}{*}{Avg} \\
\cmidrule(lr){3-4} \cmidrule(lr){5-6} \cmidrule(lr){7-8}
& & MER2023 & MER2024 & SIMS & MOSI & $S_1$ & $S_2$ & \\
\midrule
Otter & 7B & 16.41 & 14.65 & 60.57 & 54.27 & 14.64 & 4.46 & 27.50 \\
Video-LLaVA & 7B & 36.93 & 30.25 & 54.64 & 57.62 & 27.40 & 12.18 & 36.50 \\
R1-Omni & 0.5B & 59.61 & 70.92 & 58.42 & 30.17 & 32.80 & 18.85 & 45.13 \\
VideoChat2 & 7B & 33.67 & 54.50 & 69.59 & 67.23 & 34.07 & 17.78 & 46.14 \\
Video-ChatGPT & 7B & 44.86 & 46.80 & 64.43 & 57.77 & 35.29 & 19.77 & 44.82 \\
LLaMA-VID & 7B & 50.72 & 57.60 & 68.81 & 62.65 & 40.89 & 21.60 & 50.38 \\
Chat-UniVi & 7B & 57.62 & 65.67 & 67.78 & 57.62 & 43.33 & 23.90 & 52.65 \\
PandaGPT & 7B & 40.21 & 51.89 & 68.38 & 61.92 & 41.97 & 22.86 & 47.87 \\
VideoChat & 7B & 48.73 & 57.30 & 69.33 & 65.09 & 43.48 & 24.30 & 51.37 \\
mPLUG-Owl & 7B & 56.86 & 59.89 & 71.65 & 72.26 & \underline{46.28} & 27.32 & 55.71 \\
AffectGPT$^*$ & 7B & 55.23 & 59.96 & 65.65 & \textbf{76.83} & 44.98 & \underline{28.65} & 55.22 \\
\midrule
Baseline & 0.5B & 61.07 & 71.69 & 59.95 & 26.96 & 30.98 & 17.58 & 44.70 \\
Baseline & 7B & \textbf{67.39} & \textbf{76.47} & \textbf{72.90} & 71.93 & 34.09 & 18.49 & \underline{56.88} \\
Our Method & 0.5B & \underline{65.69} & \underline{72.71} & \underline{72.21} & \underline{72.59} & \textbf{55.52} & \textbf{30.48} & \textbf{61.53} \\
\bottomrule
\end{tabularx}
\end{sc}
\end{footnotesize}
\end{center}
\vskip -0.1in
\end{table*}

\textbf{Accuracy Reward.} $r_{\text{acc}}$ is the mean F1-score across five Emotion-Wheel dimensions $\mathcal{W}$, modulated by a length-penalty $P_{\ell}$ to prevent reward hacking: $r_{\text{acc}} = P_{\ell} \cdot \frac{1}{|\mathcal{W}|} \sum_{w \in \mathcal{W}} \text{F1}_w(\hat{y}, y)$.

\textbf{Protocol Consistency.} We enforce the structured sequence $\langle \texttt{hypotheses} \rangle \to \langle \texttt{think} \rangle \to \langle \texttt{answer} \rangle$ through $r_{\text{fmt}}$, which provides partial credit for valid JSON extraction. To ensure the model performs comparative adjudication rather than singular judgment, we define $r_{\text{think}}$ as a binary indicator for the presence of comparative ($\mathcal{T}_{\text{com}}$), differential ($\mathcal{T}_{\text{diff}}$), and decisive ($\mathcal{T}_{\text{dec}}$) blocks:
\begin{equation}
r_{\text{think}} = \mathbb{I}\big(\mathcal{T}_{\text{com}} \neq \emptyset \wedge \mathcal{T}_{\text{diff}} \neq \emptyset \wedge \mathcal{T}_{\text{dec}} \neq \emptyset\big).
\end{equation}

\textbf{Hierarchical Citation ($r_{\text{cite}}$).} To incentivize inter-rationale adjudication, we reward the explicit referencing of candidate hypotheses $H_k$ and the selected rationale $H^*$ within the $\langle \texttt{think} \rangle$ block:
\begin{equation}
r_{\text{cite}} = \alpha \mathbb{I}(H_k \in \mathcal{T}{\text{com}}) + \beta \mathbb{I}(H_k \in \mathcal{T}{\text{diff}}) + \gamma \mathbb{I}(H^* \in \mathcal{T}_{\text{dec}}).
\end{equation}

\textbf{Intra-trace Evidence Consistency ($r_{\text{evid}}$).} We ensure that verification-stage claims are derived from the self-declared evidence pool $V = \bigcup E_k$ defined in the proposal stage. Let $\mathcal{C}$ be the set of bracketed citations in $\langle \texttt{think} \rangle$:\begin{equation}r_{\text{evid}} = \frac{1}{|\mathcal{C}|} \sum_{c \in \mathcal{C}} \mathbb{I}\big(\mathrm{match}(c, V)\big) \quad \text{if } |\mathcal{C}| > 0 \text{ else } 0.\end{equation}The $\mathrm{match}(\cdot)$ operator employs fuzzy string matching to tolerate minor surface variations.

\textbf{Semantic Grounding ($r_{\text{sem}}$).} To ensure rationales remain rooted in multimodal reality, we align predicted evidence descriptions $P$ with ground-truth cue annotations $G$ using sentence embedding similarity $\phi(\cdot, \cdot)$:
\begin{equation}r_{\text{sem}} = \frac{1}{|P|} \sum_{p_i \in P} Q\left( \max_{g \in G} \phi(\text{emb}(p_i), \text{emb}(g)) \right),
\end{equation}where $Q(\cdot)$ discretizes continuous similarity into robust reward levels.

\textbf{Synergistic Effect.} While $r_{\text{sem}}$ and $r_{\text{evid}}$ provide fact-based grounding, $r_{\text{think}}$ and $r_{\text{cite}}$ facilitate logical synthesis. Together, this hierarchical reward system guides the policy to evolve from generating isolated ``backstory seeds" to internalizing a comprehensive reasoning trajectory that reconstructs the true affective state through the triangulation of multimodal evidence.

\begin{table}[t]
\caption{\textbf{Conflict robustness results on MER-FG.} We evaluate HyDRA against HumanOmni (Baseline) and other models across High Conflict (HCS) and Low Conflict (LCS) subsets. HyDRA maintains superior performance even under significant modal conflict. $\ddagger$: HumanOmni (7B); $^*$: AffectGPT trained on OV-MERD. Best results are \textbf{bolded}, second-best \underline{underlined}.}
\label{tab:conflict_dataset}
\begin{center}
\begin{footnotesize}
\begin{sc}
\renewcommand{\arraystretch}{1}
\newcolumntype{C}{>{\centering\arraybackslash}X}
\begin{tabularx}{\linewidth}{l C C C C}
\toprule
\multirow{2}{*}{Model} 
& \multicolumn{2}{c}{HCS} 
& \multicolumn{2}{c}{LCS} \\
\cmidrule(lr){2-3} \cmidrule(lr){4-5}
& $S_1$ & $S_2$ & $S_1$ & $S_2$ \\
\midrule
Baseline  & 30.85 & 17.24 & 31.33 & 18.50  \\
Baseline$\ddagger$ & 33.05 & 17.80 & 34.56 & 19.17  \\
PandaGPT           & 41.15 & 22.15 & 45.43 & 25.59  \\
AffectGPT$^*$      & \underline{43.63} & \underline{27.89} & \underline{48.82} & \underline{29.76} \\
\midrule
HyDRA         & \textbf{54.78} & \textbf{30.56} & \textbf{55.51} & \textbf{30.13} \\
\bottomrule
\end{tabularx}
\end{sc}
\end{footnotesize}
\end{center}
\vskip -0.1in
\end{table}

\section{Experiments}

\begin{figure*}[ht]  
\vskip 0.2in  
\begin{center}    
\centerline{\includegraphics[width=0.92\textwidth]{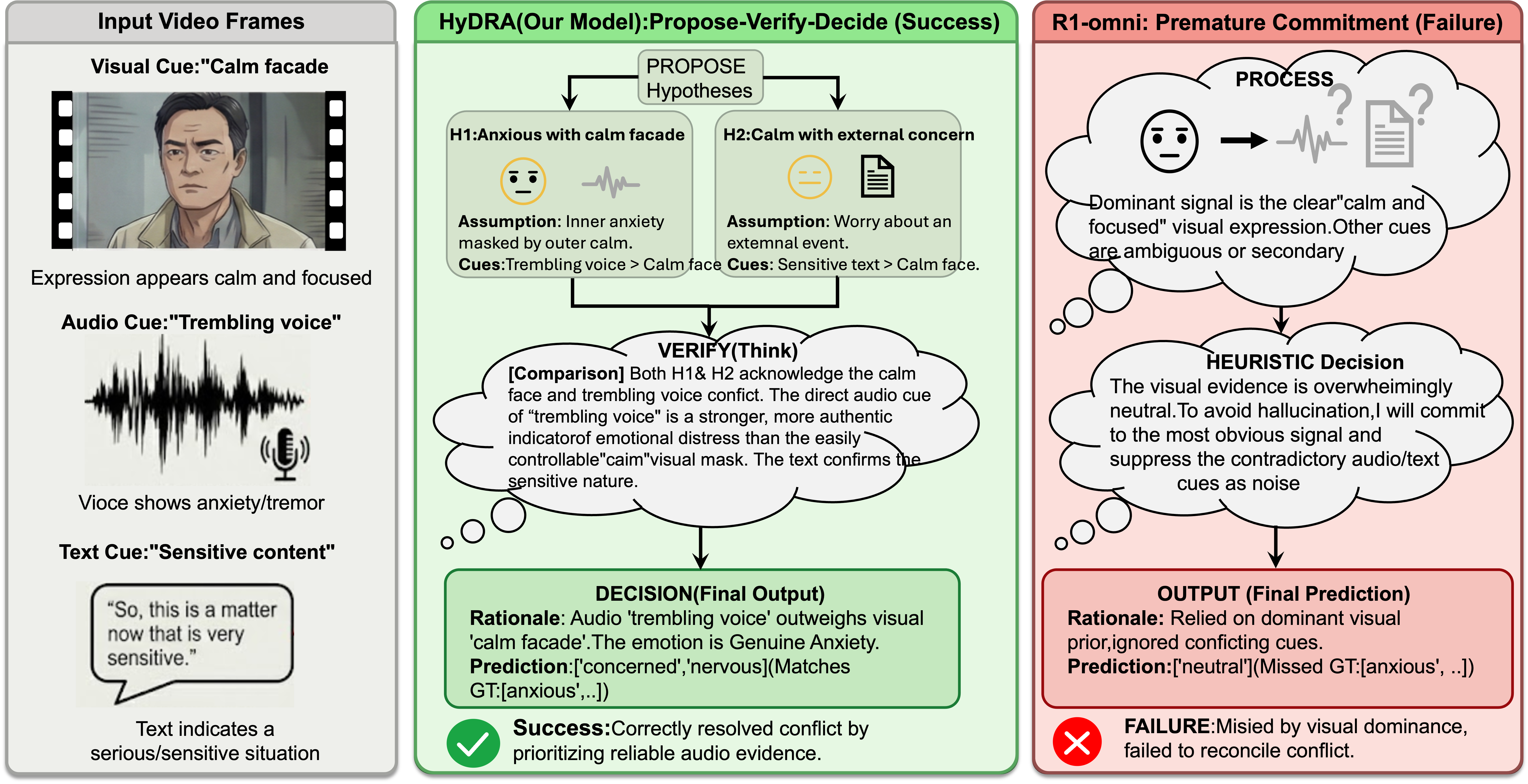}}    
\caption{      
\textbf{Left}: The original input (visual/audio/text). Visual priors contradict subtle audio/textual cues. \textbf{Middle (Success)}: HyDRA resolves the conflict via the Propose–Verify–Decide protocol. \textbf{Right (Failure)}: R1-omni commits prematurely to salient visual signals. Due to the presence of real individuals in the original videos, personal identifiable information has been removed and processed via visualization to address copyright and privacy concerns.
}    
\label{fig:Case_Study}  
\end{center}
\end{figure*}

\subsection{Datasets \& Evaluation Metrics}

\textbf{Cold-start Subset.} To establish a robust initial capability for open-vocabulary reasoning, we utilize \textbf{OV-MERD} \cite{lian2024ov} and 242 expert-verified samples from \textbf{MERCaption+} \cite{lian2025affectgpt}, ensuring high semantic density for the cold-start phase.

\textbf{GRPO RL Subset.} For the RL stage, we curate 12,000 samples from MERCaption+, manually filtered by audio-visual clarity and alignment accuracy. Both subsets are partitioned into a unified \textbf{Observation Graph (ObsG)} format; the full preparation process is detailed in Appendix~\ref{app:data_preparation}. We adopt HumanOmni-0.5B \cite{zhao2025humanomni} as our backbone, noting its architectural differences from the 7B version in the Appendix~\ref{app:baselines}.

\noindent\textbf{Benchmarks.} We evaluate HyDRA across two primary dimensions: (1) \textit{General Emotion Recognition}: We assess the model's performance on both sentiment-level intensity and discrete emotional categories. Specifically, we utilize CMU-MOSI~\cite{zadeh2017tensor} and CH-SIMS~\cite{yu2020ch} to evaluate sentiment understanding, while MER2023~\cite{10.1145/3581783.3612836} and MER2024~\cite{lian2024mer} are employed to test the recognition of basic emotional categories.
(2) \textit{Open-Vocabulary Fine-Grained (OV-FG)}: We adopt the \textbf{MER-FG} benchmark \cite{lian_ovmer} (a curated subset of MERCaption+), which provides a high-density label space and serves as a rigorous stress test for open-set affective reasoning and deductive precision.

Crucially, although MERCaption+ is utilized across different stages, we ensure that the three subsets—the 200 SFT samples, the 12,000 RL samples, and the OV-FG evaluation set—are \textbf{strictly disjoint}. This isolation guarantees that the evaluation reflects true zero-shot or out-of-distribution performance without any overlap with the training data.

\noindent\textbf{Evaluation Metrics.} For the Sentiment and Basic emotion tasks, we strictly follow the protocols in \cite{lian2025affectgpt}, adopting ACC. Following the MER-FG protocol \cite{lian2025mer}, we report the average F1-score across two granularities on OV-FG: $S_1$ (coarse-grained) and $S_2$ (fine-grained). To handle open-vocabulary synonyms and varying label densities, both metrics apply hierarchical grouping via multiple emotion wheels to map predictions into unified taxonomic levels before calculating the sample-wise F1-score. 

Unless otherwise noted, all experimental results are reported in percentages (\%). Full details regarding benchmark statistics, baseline descriptions, metric specifications, and implementation configurations are provided in Appendix~\ref{app:datasets}, \ref{app:baselines}, \ref{app:ew_metric}, and \ref{app:experimental_setup}, respectively.

\subsection{Main Result}

Table~\ref{tab:metrics_track2} shows our method achieves the best overall performance, despite using a 0.5B backbone, reaching an average score of 61.53 across all six evaluations. This indicates that the improvement is primarily driven by the proposed Propose–Verify–Decide inference, rather than model scale.

\textbf{Strength on open-vocabulary fine-grained emotion.} The largest gains appear on OV-FG, where our method ranks first on both granularities. In particular, it substantially improves the coarse-grained score and also sets a new best fine-grained score. This aligns with our motivation: when label space is open and cues are underspecified, a single prior-driven interpretation is brittle, while multi-hypothesis proposal followed by evidence-constrained verification better reconciles complementary cues.

\textbf{Robustness on basic emotion and sentiment.} Beyond OV-FG, our method remains consistently strong on Basic and Sentiment tasks, achieving second-best performance across MER2023, MER2024, SIMS, and MOSI. Notably, it stays competitive with 7B baselines on sentiment while preserving strong basic-emotion recognition, suggesting that enforcing evidence-closed reasoning does not trade off general affect recognition capability for fine-grained gains.

\newcommand{\on}{\textbullet}   
\newcommand{\off}{\textopenbullet} 
\begin{table}[t]
  \caption{Ablation study of GRPO reward components on MER-FG. Symbols $\bullet$ and $\circ$ indicate enabled and disabled rewards, respectively. For.: Format constraints; Acc.: Accuracy reward; Cit.: Hierarchical citation; Sem.: Semantic Grounding. \textbf{Bold}: Best; \underline{Underline}: Second-best.}
  \label{tab:ablation_reward}
  \begin{center}
    \begin{small}
      \begin{sc}
        \begin{tabularx}{\columnwidth}{cccc ccc}
          \toprule
          \multirow{2}{*}{For.} & 
          \multirow{2}{*}{Acc.} & 
          \multirow{2}{*}{Cit.} & 
          \multirow{2}{*}{Sem.} & 
          \multicolumn{3}{c}{OV-FG} \\
          \cmidrule(l){5-7}
          & & & & $S_1$ & $S_2$ & Avg  \\
          \midrule
            \on  & \off & \off & \off & 47.93 & 25.67 & 36.80 \\
            \off & \on  & \off & \off & 49.27 & 25.73 & 37.50 \\
            \off & \off & \on  & \off & 47.30 & 25.85 & 36.57 \\
            \off & \off & \off & \on  & 49.79 & 26.00 & 37.90 \\
            \midrule
            \on & \off  & \on  & \on  & 50.46 & 26.35 & 38.40 \\
            \on  & \on  & \off & \on  & \underline{52.65} & \underline{27.88} & \underline{40.26} \\
            \on  & \on  & \on  & \off & 52.42 & 26.08 & 39.25 \\
            \on  & \on  & \on  & \on  & \textbf{55.52} & \textbf{30.48} & \textbf{43.00} \\
          \bottomrule
        \end{tabularx}
      \end{sc}
    \end{small}
  \end{center}
  \vskip -0.1in
\end{table}

\textbf{Conflict robustness.} Table~\ref{tab:conflict_dataset} stratifies the test set by cross-modal conflict. All baselines drop noticeably on HCS compared to LCS, confirming their fragility under contradictory cues. HyDRA stays best on both subsets and degrades the least, indicating that multi-path adjudication mitigates premature commitment when modalities disagree. Compared to AffectGPT, HyDRA improves $S_1$ by +11.15 on HCS versus +6.69 on LCS, showing larger relative gains in conflict-heavy cases.

\begin{table}[t]
  \caption{\textbf{Impact of hypothesis cardinality ($K$) on MER-FG.} "No-hypotheses" denotes the standard linear reasoning (R1-style) without the Propose--Verify--Decide protocol. $K=2$ achieves the optimal balance between analytical diversity and evidential focus. All variants use the same 0.5B backbone. \textbf{Bold}: Best; \underline{Underline}: Second-best.}
  \label{tab:proof_hypo}
  \begin{center}
    \begin{small}
      \begin{sc}
        \begin{tabularx}{\columnwidth}{X ccc}
          \toprule
          \multirow{2}{*}{Model} & \multicolumn{3}{c}{OV-FG} \\
          \cmidrule(l){2-4}
          & $S_1$ & $S_2$ & Avg\\
          \midrule
          Baseline      & 30.98 & 17.58 & 24.28 \\
          No-hypotheses & 52.14 & \underline{26.88} & 39.51 \\
          1-hypothesis & 49.24 & 25.50 & 37.37\\
          3-hypotheses & \underline{54.48} & 26.05 & \underline{40.26} \\
          4-hypotheses & 53.51 & 26.28 & 39.89 \\
          \midrule
          HyDRA     & \textbf{55.52} & \textbf{30.48} & \textbf{43.00} \\
          \bottomrule
        \end{tabularx}
      \end{sc}
    \end{small}
  \end{center}
  \vskip -0.1in
\end{table}

Figure~\ref{fig:Case_Study} illustrates a representative scenario where HyDRA effectively resolves modality conflicts that mislead standard models. In this case, the subject maintains a ``calm and focused" facial expression , while the audio and text cues signify intense internal distress. Unlike the R1-omni, which suffers from premature commitment to the salient visual signal, HyDRA avoids an immediate decision. Through the Propose–Verify–Decide protocol, it generates competing hypotheses and correctly adjudicates that the subtle audio cue is a more authentic indicator of affect than the controlled visual mask.
By reconciling these contradictory signals, HyDRA successfully identifies the state as Genuine Anxiety, matching the ground truth while providing a transparent, evidence-grounded reasoning trace.

A detailed discussion regarding the scope and limitations of our proposed method is provided in Appendix~\ref{app:limitations}.

\section{Ablation Study}
This section provides a systematic decomposition of HyDRA to evaluate the contribution of its core components: reward design, reasoning structure, and training paradigms. Unless otherwise specified, all variants are evaluated on the OV-FG using its standard metrics. In every ablation configuration, the RL stage follows the same training schedule as the formal experiments. The results reported here reflect the alignment behavior across different design choices, providing a direct comparison of the impact of each component on the model's performance.

\subsection{Ablating GRPO Rewards}

We study how each reward component contributes to RL alignment under HyDRA.  We evaluate three settings: enabling all rewards (default), \emph{keep-one-only}, and \emph{leave-one-out}, as summarized in Table~\ref{tab:ablation_reward}.

\textbf{Keep-one-only.}
When only one reward is enabled, Accuracy reward and Evidence selection reward yield the strongest performance. This is expected because both provide task-aligned content supervision: one targets the final prediction quality, while the other constrains evidence quality via semantic matching. In contrast, Format rewards and Hierarchical evidence citation rewards offer limited gains in isolation, suggesting that structural compliance alone is insufficient to ensure correct open-vocabulary emotion inference.

\textbf{Leave-one-out.}
In multi-reward RL, removing the Accuracy reward causes a clear degradation, indicating that it remains the dominant task-aligned learning signal, which employs $P_\ell$ to mitigate reward hacking (see Appendix~\ref{app:length_penalty}). Conversely, removing  Hierarchical evidence citation rewards yields the strongest performance among leave-one-out variants. 
In our final training configuration, Format rewards are consistently enabled as a foundational constraint, rather than a primary variable in the leave-one-out analysis. This decision is made because removing format rewards in RL yields high numerical scores , yet it incurs the risk of long-term structural degradation. By fixing format rewards as a constant, we ensure the model maintains strict structural integrity, while simultaneously isolating the impacts of content-driven rewards.

\begin{table}[t]
  \caption{\textbf{Comparison of training paradigms on MER-FG. }Cold-start and GRPO RL Subsets are used for initial and RL stages. $\text{SFT}_\text{full}$, PPO, and HyDRA share the same data budget. Results show that RL (especially our GRPO-based HyDRA) significantly outperforms SFT scaling. \textbf{Bold}: Best; \underline{Underline}: Second-best.}
  \label{tab:proof_rl}
  \begin{center}
    \begin{small}
      \begin{sc}
        \begin{tabularx}{\columnwidth}{X ccc}
          \toprule
          \multirow{2}{*}{Model} & \multicolumn{3}{c}{OV-FG} \\
          \cmidrule(l){2-4}
          & $S_1$ & $S_2$ & Avg \\
          \midrule
          Prompt-only   & 23.26 & 12.34 & 17.80 \\
          Baseline      & 30.98 & 17.58 & 24.28 \\
          $\text{SFT}_\text{cold}$          & 45.91 & 24.22 & 35.07 \\
          PPO Training       & \underline{51.09} & \underline{26.93} & \underline{39.01} \\
           $\text{SFT}_\text{full}$  & 49.60 & 25.31 & 37.46 \\
          \midrule
          HyDRA           & \textbf{55.52} & \textbf{30.48} & \textbf{43.00} \\
          \bottomrule
        \end{tabularx}
      \end{sc}
    \end{small}
  \end{center}
  \vskip -0.1in
\end{table}

\subsection{Impact of Hypothesis Cardinality on Reasoning}
Table~\ref{tab:proof_hypo} evaluates hypothesis cardinality $K$. Every Propose--Verify--Decide variant significantly outperforms the baseline, confirming that the structural benefit of the Propose–Verify–Decide protocol is fundamental. However, the ``No-hypotheses" variant—representing a linear reasoning paradigm akin to the DeepSeek-R1 style—remains suboptimal. While linear traces allow for logical depth, they lack the explicit divergent-then-convergent mechanism necessary to reconcile the ambiguous and often contradictory cues inherent in affective reasoning.

We identify a performance ``sweet spot" at $K=2$, which balances analytical diversity with token efficiency. Interestingly, the $K=1$ variant performs worse than the linear ``No-hypotheses" baseline, revealing a confirmation bias trap: forced upfront commitment causes the verification phase to anchor on justifying the initial guess rather than rather than performing the ``cross-examination" described in Sec.~\ref{sec:protocol}. $K=2$ introduces sufficient ``cognitive friction" to compel genuine adjudication without exceeding the informational capacity of typical 2–5 second video clips.

Beyond $K=2$, returns diminish due to the sparse evidence in short-form content. $K=3$ often leads to semantic redundancy where candidates overlap significantly, while $K=4$ triggers ``over-interpretation." When forced to generate excessive diversity from limited cues, the model may hallucinate situational dynamics to satisfy the cardinality constraint, introducing noise. Notably, while $S_2$ shows slight resilience at $K=4$ by occasionally capturing rare fine-grained labels, the overall reliability and latency trade-offs favor the more robust $K=2$ configuration.

\subsection{Training Paradigm Boundary}

Table~\ref{tab:proof_rl} investigates whether HyDRA's reasoning capability stems from parameter internalization or mere data scaling. \textbf{Necessity of Adaptation. }The ``Prompt-only" approach, which applies our Propose--Verify--Decide protocol to the frozen backbone, performs significantly worse than the baseline. This confirms that the multi-step reasoning logic is not a superficial prompting trick but a complex deductive behavior that must be internalized through systematic parameter updates. \textbf{RL vs. SFT Efficiency.} While $\text{SFT}_\text{cold}$ establishes a solid foundation for open-vocabulary tasks, expanding the supervised data to $\text{SFT}_\text{full}$ yields only marginal improvements. Notably, both RL-based paradigms—PPO and HyDRA—outperform $\text{SFT}_\text{full}$ using the same data budget. This gap demonstrates that for fine-grained affective reasoning, reinforcement learning is more sample-efficient than density-driven supervised learning, as it encourages the model to explore and self-correct reasoning paths. \textbf{Superiority of HyDRA.} Among RL variants, HyDRA further surpasses PPO by 3.99\% in average score, achieving the peak performance. This suggests that our GRPO framework, tailored with evidence-constrained rewards, more effectively stabilizes the ``Propose–Verify–Decide" trajectory, preventing the model from regressing to biased heuristics.

\section{Conclusion}
We introduced \textbf{HyDRA}, which operationalizes Open-Vocabulary MER as a \textit{Propose–Verify–Decide} protocol.
HyDRA repurposes MLLM generative priors to propose multiple competing latent-context hypotheses, and then performs evidence-constrained adjudication to select the emotion set that best reconciles equivocal and potentially conflicting multimodal cues. 
To make this protocol a learned capability rather than a prompting heuristic, we further optimized the model with \textbf{GRPO} and \textit{hierarchical reward shaping}, explicitly aligning intermediate reasoning trajectories with final task performance and enforcing \textit{evidential closure}. 
Across systematic evaluations, HyDRA consistently outperformed strong baselines on diverse OV-MER benchmarks, with particularly pronounced gains under ambiguity and cross-modal conflict, and it notably surpassed the performance of an zero-shot 7B baseline. 
Beyond metric gains, HyDRA produces interpretable and diagnostic evidence traces, enabling principled analysis of when and why the model commits to a prediction.
We hope this work encourages the community to treat OV-MER as a hybrid abductive--deductive inference problem and to build learning objectives that reward verifiable, evidence-grounded reasoning.


\section*{Impact Statement}
This work enhances the reliability and interpretability of Multimodal Emotion Recognition by mitigating model bias in ambiguous scenarios. While our framework enables more empathetic and transparent AI for applications in mental health and human-computer interaction, we recognize the inherent risks of emotional profiling and privacy infringement associated with affective computing. We advocate for the deployment of these technologies exclusively within ethical frameworks that prioritize user consent and prohibit unauthorized surveillance.
\nocite{langley00}

\bibliographystyle{icml2026}

\begin{thebibliography}{61}
\providecommand{\natexlab}[1]{#1}
\providecommand{\url}[1]{\texttt{#1}}
\expandafter\ifx\csname urlstyle\endcsname\relax
  \providecommand{\doi}[1]{doi: #1}\else
  \providecommand{\doi}{doi: \begingroup \urlstyle{rm}\Url}\fi

\bibitem[Bai et~al.(2024)Bai, Wang, Xiao, He, Han, Zhang, and Shou]{bai2024hallucination}
Bai, Z., Wang, P., Xiao, T., He, T., Han, Z., Zhang, Z., and Shou, M.~Z.
\newblock Hallucination of multimodal large language models: A survey.
\newblock \emph{arXiv preprint arXiv:2404.18930}, 2024.

\bibitem[Barrett et~al.(2019)Barrett, Adolphs, Marsella, Martinez, and Pollak]{doi:10.1177/1529100619832930}
Barrett, L.~F., Adolphs, R., Marsella, S., Martinez, A.~M., and Pollak, S.~D.
\newblock Emotional expressions reconsidered: Challenges to inferring emotion from human facial movements.
\newblock \emph{Psychological Science in the Public Interest}, 20\penalty0 (1):\penalty0 1--68, 2019.
\newblock \doi{10.1177/1529100619832930}.
\newblock PMID: 31313636.

\bibitem[Bhattacharyya \& Wang(2025)Bhattacharyya and Wang]{bhattacharyya-wang-2025-evaluating}
Bhattacharyya, S. and Wang, J.~Z.
\newblock Evaluating vision-language models for emotion recognition.
\newblock In Chiruzzo, L., Ritter, A., and Wang, L. (eds.), \emph{Findings of the Association for Computational Linguistics: NAACL 2025}, pp.\  1798--1820, Albuquerque, New Mexico, April 2025. Association for Computational Linguistics.
\newblock ISBN 979-8-89176-195-7.
\newblock \doi{10.18653/v1/2025.findings-naacl.97}.

\bibitem[Chang et~al.(2026)Chang, Wang, Guo, Nan, Yao, and Zhou]{chang2026abductivemllmboostingvisualabductive}
Chang, B., Wang, Q., Guo, X., Nan, Z., Yao, Y., and Zhou, T.
\newblock Abductivemllm: Boosting visual abductive reasoning within mllms, 2026.

\bibitem[Dang et~al.(2025)Dang, Song, Xiao, Wang, Peng, Li, Yang, Wang, and Chua]{dang2025mupamultipathagenticreasoning}
Dang, J., Song, H., Xiao, J., Wang, B., Peng, H., Li, H., Yang, X., Wang, M., and Chua, T.-S.
\newblock Mupa: Towards multi-path agentic reasoning for grounded video question answering, 2025.

\bibitem[Dhuliawala et~al.(2024)Dhuliawala, Komeili, Xu, Raileanu, Li, Celikyilmaz, and Weston]{dhuliawala2024chain}
Dhuliawala, S., Komeili, M., Xu, J., Raileanu, R., Li, X., Celikyilmaz, A., and Weston, J.
\newblock Chain-of-verification reduces hallucination in large language models.
\newblock In \emph{Findings of the association for computational linguistics: ACL 2024}, pp.\  3563--3578, 2024.

\bibitem[Ge et~al.(2024)Ge, Tang, and Li]{ge2024video}
Ge, M., Tang, D., and Li, M.
\newblock Video emotion open-vocabulary recognition based on multimodal large language model.
\newblock \emph{arXiv preprint arXiv:2408.11286}, 2024.

\bibitem[Geirhos et~al.(2020)Geirhos, Jacobsen, Michaelis, Zemel, Brendel, Bethge, and Wichmann]{geirhos2020shortcut}
Geirhos, R., Jacobsen, J.-H., Michaelis, C., Zemel, R., Brendel, W., Bethge, M., and Wichmann, F.~A.
\newblock Shortcut learning in deep neural networks.
\newblock \emph{Nature Machine Intelligence}, 2\penalty0 (11):\penalty0 665--673, 2020.

\bibitem[Goyal et~al.(2017)Goyal, Khot, Summers-Stay, Batra, and Parikh]{goyal2017making}
Goyal, Y., Khot, T., Summers-Stay, D., Batra, D., and Parikh, D.
\newblock Making the v in vqa matter: Elevating the role of image understanding in visual question answering.
\newblock In \emph{Proceedings of the IEEE conference on computer vision and pattern recognition}, pp.\  6904--6913, 2017.

\bibitem[Guerdelli et~al.(2023)Guerdelli, Ferrari, Cardia~Neto, Berretti, Barhoumi, and Del~Bimbo]{guerdelli2023towards}
Guerdelli, H., Ferrari, C., Cardia~Neto, J.~B., Berretti, S., Barhoumi, W., and Del~Bimbo, A.
\newblock Towards a better understanding of human emotions: Challenges of dataset labeling.
\newblock In \emph{International Conference on Image Analysis and Processing}, pp.\  242--254. Springer, 2023.

\bibitem[Guo et~al.(2025)Guo, Yang, Zhang, Song, Wang, Zhu, Xu, Zhang, Ma, Bi, et~al.]{guo2025deepseek}
Guo, D., Yang, D., Zhang, H., Song, J., Wang, P., Zhu, Q., Xu, R., Zhang, R., Ma, S., Bi, X., et~al.
\newblock Deepseek-r1 incentivizes reasoning in llms through reinforcement learning.
\newblock \emph{Nature}, 645\penalty0 (8081):\penalty0 633--638, 2025.

\bibitem[Han et~al.(2025{\natexlab{a}})Han, Zhu, Xu, Song, and Yang]{10.1145/3746027.3754856}
Han, Z., Zhu, B., Xu, Y., Song, P., and Yang, X.
\newblock Benchmarking and bridging emotion conflicts for multimodal emotion reasoning.
\newblock In \emph{Proceedings of the 33rd ACM International Conference on Multimedia}, MM '25, pp.\  5528–5537, New York, NY, USA, 2025{\natexlab{a}}. Association for Computing Machinery.
\newblock ISBN 9798400720352.
\newblock \doi{10.1145/3746027.3754856}.

\bibitem[Han et~al.(2025{\natexlab{b}})Han, Zhu, Xu, Song, and Yang]{han2025benchmarking}
Han, Z., Zhu, B., Xu, Y., Song, P., and Yang, X.
\newblock Benchmarking and bridging emotion conflicts for multimodal emotion reasoning.
\newblock In \emph{Proceedings of the 33rd ACM International Conference on Multimedia}, pp.\  5528--5537, 2025{\natexlab{b}}.

\bibitem[Huang et~al.(2024)Huang, Dong, Zhang, Wang, He, Wang, Lin, Zhang, and Yu]{huang2024opera}
Huang, Q., Dong, X., Zhang, P., Wang, B., He, C., Wang, J., Lin, D., Zhang, W., and Yu, N.
\newblock Opera: Alleviating hallucination in multi-modal large language models via over-trust penalty and retrospection-allocation.
\newblock In \emph{Proceedings of the IEEE/CVF Conference on Computer Vision and Pattern Recognition}, pp.\  13418--13427, 2024.

\bibitem[Jin et~al.(2024)Jin, Takanobu, Zhang, Cao, and Yuan]{jin2024chat}
Jin, P., Takanobu, R., Zhang, W., Cao, X., and Yuan, L.
\newblock Chat-univi: Unified visual representation empowers large language models with image and video understanding.
\newblock In \emph{Proceedings of the IEEE/CVF Conference on Computer Vision and Pattern Recognition}, pp.\  13700--13710, 2024.

\bibitem[Kahneman(2011)]{kahneman2011thinking}
Kahneman, D.
\newblock \emph{Thinking, Fast and Slow}.
\newblock Farrar, Straus and Giroux, New York, 2011.

\bibitem[Lei et~al.(2024)Lei, Yang, Chen, Chen, Zhai, and Zhang]{lei2024large}
Lei, Y., Yang, D., Chen, Z., Chen, J., Zhai, P., and Zhang, L.
\newblock Large vision-language models as emotion recognizers in context awareness.
\newblock \emph{arXiv preprint arXiv:2407.11300}, 2024.

\bibitem[Leng et~al.(2024)Leng, Zhang, Chen, Li, Lu, Miao, and Bing]{leng2024mitigating}
Leng, S., Zhang, H., Chen, G., Li, X., Lu, S., Miao, C., and Bing, L.
\newblock Mitigating object hallucinations in large vision-language models through visual contrastive decoding.
\newblock In \emph{Proceedings of the IEEE/CVF Conference on Computer Vision and Pattern Recognition}, pp.\  13872--13882, 2024.

\bibitem[Li et~al.(2025{\natexlab{a}})Li, Zhang, Chen, Wang, Pu, Cahyono, Yang, Li, and Liu]{li2025otter}
Li, B., Zhang, Y., Chen, L., Wang, J., Pu, F., Cahyono, J.~A., Yang, J., Li, C., and Liu, Z.
\newblock Otter: A multi-modal model with in-context instruction tuning.
\newblock \emph{IEEE Transactions on Pattern Analysis and Machine Intelligence}, 2025{\natexlab{a}}.

\bibitem[Li et~al.(2024{\natexlab{a}})Li, Zhang, Zhang, Zhang, Li, Li, Ma, and Li]{li2024llavanextinterleavetacklingmultiimagevideo}
Li, F., Zhang, R., Zhang, H., Zhang, Y., Li, B., Li, W., Ma, Z., and Li, C.
\newblock Llava-next-interleave: Tackling multi-image, video, and 3d in large multimodal models, 2024{\natexlab{a}}.

\bibitem[Li et~al.(2025{\natexlab{b}})Li, He, Wang, Li, Wang, Luo, Wang, Wang, and Qiao]{li2025videochat}
Li, K., He, Y., Wang, Y., Li, Y., Wang, W., Luo, P., Wang, Y., Wang, L., and Qiao, Y.
\newblock Videochat: Chat-centric video understanding.
\newblock \emph{Science China Information Sciences}, 68\penalty0 (10):\penalty0 200102, 2025{\natexlab{b}}.

\bibitem[Li et~al.(2023)Li, Du, Zhou, Wang, Zhao, and Wen]{li-etal-2023-evaluating}
Li, Y., Du, Y., Zhou, K., Wang, J., Zhao, X., and Wen, J.-R.
\newblock Evaluating object hallucination in large vision-language models.
\newblock In Bouamor, H., Pino, J., and Bali, K. (eds.), \emph{Proceedings of the 2023 Conference on Empirical Methods in Natural Language Processing}, pp.\  292--305, Singapore, December 2023. Association for Computational Linguistics.
\newblock \doi{10.18653/v1/2023.emnlp-main.20}.

\bibitem[Li et~al.(2024{\natexlab{b}})Li, Wang, and Jia]{li2024llama}
Li, Y., Wang, C., and Jia, J.
\newblock Llama-vid: An image is worth 2 tokens in large language models.
\newblock In \emph{European Conference on Computer Vision}, pp.\  323--340. Springer, 2024{\natexlab{b}}.

\bibitem[Li et~al.(2025{\natexlab{c}})Li, Wu, Shi, Qin, Du, Liu, Zhou, Manocha, and Boyd-Graber]{li2025videohallu}
Li, Z., Wu, X., Shi, G., Qin, Y., Du, H., Liu, F., Zhou, T., Manocha, D., and Boyd-Graber, J.~L.
\newblock Videohallu: Evaluating and mitigating multi-modal hallucinations on synthetic video understanding.
\newblock \emph{arXiv preprint arXiv:2505.01481}, 2025{\natexlab{c}}.

\bibitem[Lian et~al.(2023{\natexlab{a}})Lian, Sun, Sun, Chen, Xu, Wang, Xu, He, Li, Zhao, Liu, Liu, Yi, Wang, Cambria, Zhao, Schuller, and Tao]{10.1145/3581783.3612836}
Lian, Z., Sun, H., Sun, L., Chen, K., Xu, M., Wang, K., Xu, K., He, Y., Li, Y., Zhao, J., Liu, Y., Liu, B., Yi, J., Wang, M., Cambria, E., Zhao, G., Schuller, B.~W., and Tao, J.
\newblock Mer 2023: Multi-label learning, modality robustness, and semi-supervised learning.
\newblock In \emph{Proceedings of the 31st ACM International Conference on Multimedia}, MM '23, pp.\  9610–9614, New York, NY, USA, 2023{\natexlab{a}}. Association for Computing Machinery.
\newblock ISBN 9798400701085.
\newblock \doi{10.1145/3581783.3612836}.

\bibitem[Lian et~al.(2023{\natexlab{b}})Lian, Sun, Sun, Gu, Wen, Zhang, Chen, Xu, Xu, Chen, et~al.]{lian2023explainable}
Lian, Z., Sun, H., Sun, L., Gu, H., Wen, Z., Zhang, S., Chen, S., Xu, M., Xu, K., Chen, K., et~al.
\newblock Explainable multimodal emotion recognition.
\newblock \emph{arXiv preprint arXiv:2306.15401}, 2023{\natexlab{b}}.

\bibitem[Lian et~al.(2024{\natexlab{a}})Lian, Sun, Sun, Chen, Chen, Gu, Wen, Chen, Zhang, Yao, et~al.]{lian2024ov}
Lian, Z., Sun, H., Sun, L., Chen, H., Chen, L., Gu, H., Wen, Z., Chen, S., Zhang, S., Yao, H., et~al.
\newblock Ov-mer: Towards open-vocabulary multimodal emotion recognition.
\newblock \emph{arXiv preprint arXiv:2410.01495}, 2024{\natexlab{a}}.

\bibitem[Lian et~al.(2024{\natexlab{b}})Lian, Sun, Sun, Wen, Zhang, Chen, Gu, Zhao, Ma, Chen, et~al.]{lian2024mer}
Lian, Z., Sun, H., Sun, L., Wen, Z., Zhang, S., Chen, S., Gu, H., Zhao, J., Ma, Z., Chen, X., et~al.
\newblock Mer 2024: Semi-supervised learning, noise robustness, and open-vocabulary multimodal emotion recognition.
\newblock In \emph{Proceedings of the 2nd International Workshop on Multimodal and Responsible Affective Computing}, pp.\  41--48, 2024{\natexlab{b}}.

\bibitem[Lian et~al.(2025{\natexlab{a}})Lian, Chen, Chen, Sun, Sun, Ren, Cheng, Liu, Liu, Peng, Yi, and Tao]{lian2025affectgpt}
Lian, Z., Chen, H., Chen, L., Sun, H., Sun, L., Ren, Y., Cheng, Z., Liu, B., Liu, R., Peng, X., Yi, J., and Tao, J.
\newblock Affect{GPT}: A new dataset, model, and benchmark for emotion understanding with multimodal large language models.
\newblock In \emph{Forty-second International Conference on Machine Learning}, 2025{\natexlab{a}}.

\bibitem[Lian et~al.(2025{\natexlab{b}})Lian, Liu, Xu, Liu, Liu, Zhang, Liu, Li, Cheng, Zuo, et~al.]{lian2025mer}
Lian, Z., Liu, R., Xu, K., Liu, B., Liu, X., Zhang, Y., Liu, X., Li, Y., Cheng, Z., Zuo, H., et~al.
\newblock Mer 2025: When affective computing meets large language models.
\newblock In \emph{Proceedings of the 33rd ACM International Conference on Multimedia}, pp.\  13837--13842, 2025{\natexlab{b}}.

\bibitem[Lian et~al.(2024{\natexlab{c}})]{lian_ovmer}
Lian, Z. et~al.
\newblock Ov-mer: Towards open-vocabulary multimodal emotion recognition.
\newblock \emph{arXiv preprint arXiv:2410.01495}, 2024{\natexlab{c}}.

\bibitem[Lin et~al.(2024)Lin, Ye, Zhu, Cui, Ning, Jin, and Yuan]{lin2024video}
Lin, B., Ye, Y., Zhu, B., Cui, J., Ning, M., Jin, P., and Yuan, L.
\newblock Video-llava: Learning united visual representation by alignment before projection.
\newblock In \emph{Proceedings of the 2024 conference on empirical methods in natural language processing}, pp.\  5971--5984, 2024.

\bibitem[Liu et~al.(2023)Liu, Li, Wu, and Lee]{liu2023visual}
Liu, H., Li, C., Wu, Q., and Lee, Y.~J.
\newblock Visual instruction tuning.
\newblock \emph{Advances in neural information processing systems}, 36:\penalty0 34892--34916, 2023.

\bibitem[Maaz et~al.(2024)Maaz, Rasheed, Khan, and Khan]{maaz2024video}
Maaz, M., Rasheed, H., Khan, S., and Khan, F.
\newblock Video-chatgpt: Towards detailed video understanding via large vision and language models.
\newblock In \emph{Proceedings of the 62nd Annual Meeting of the Association for Computational Linguistics (Volume 1: Long Papers)}, pp.\  12585--12602, 2024.

\bibitem[Mistretta et~al.(2025)Mistretta, Baldrati, Agnolucci, Bertini, and Bagdanov]{mistretta2025cross}
Mistretta, M., Baldrati, A., Agnolucci, L., Bertini, M., and Bagdanov, A.~D.
\newblock Cross the gap: Exposing the intra-modal misalignment in clip via modality inversion.
\newblock \emph{arXiv preprint arXiv:2502.04263}, 2025.

\bibitem[Mohsin et~al.(2025)Mohsin, Umer, Bilal, Memon, Qadir, Bhattacharya, Rizwan, Gorle, Kazmi, Mohsin, et~al.]{mohsin2025fundamental}
Mohsin, M.~A., Umer, M., Bilal, A., Memon, Z., Qadir, M.~I., Bhattacharya, S., Rizwan, H., Gorle, A.~R., Kazmi, M.~Z., Mohsin, A., et~al.
\newblock On the fundamental limits of llms at scale.
\newblock \emph{arXiv preprint arXiv:2511.12869}, 2025.

\bibitem[Poria et~al.(2017)Poria, Cambria, Bajpai, and Hussain]{poria2017review}
Poria, S., Cambria, E., Bajpai, R., and Hussain, A.
\newblock A review of affective computing: From unimodal analysis to multimodal fusion.
\newblock \emph{Information fusion}, 37:\penalty0 98--125, 2017.

\bibitem[Qian et~al.(2025)Qian, Wan, Jia, Yang, Zhao, and Gan]{qian2025prismbenchbenchmarkpuzzlebasedvisual}
Qian, Y., Wan, C., Jia, C., Yang, Y., Zhao, Q., and Gan, Z.
\newblock Prism-bench: A benchmark of puzzle-based visual tasks with cot error detection, 2025.

\bibitem[Rha et~al.(2025)Rha, Yeo, Won, Park, and Ro]{rha2025learningattendfirstmodalityimportanceguided}
Rha, H., Yeo, J.~H., Won, J., Park, S.~J., and Ro, Y.~M.
\newblock Learning what to attend first: Modality-importance-guided reasoning for reliable multimodal emotion understanding, 2025.

\bibitem[Shao et~al.(2024)Shao, Wang, Zhu, Xu, Song, Bi, Zhang, Zhang, Li, Wu, and Guo]{shao2024deepseekmathpushinglimitsmathematical}
Shao, Z., Wang, P., Zhu, Q., Xu, R., Song, J., Bi, X., Zhang, H., Zhang, M., Li, Y.~K., Wu, Y., and Guo, D.
\newblock Deepseekmath: Pushing the limits of mathematical reasoning in open language models, 2024.

\bibitem[Su et~al.(2023)Su, Lan, Li, Xu, Wang, and Cai]{su2023pandagptmodelinstructionfollow}
Su, Y., Lan, T., Li, H., Xu, J., Wang, Y., and Cai, D.
\newblock Pandagpt: One model to instruction-follow them all, 2023.

\bibitem[Sun \& Saparov(2025)Sun and Saparov]{sun2025languagemodelsfollowoccams}
Sun, Y. and Saparov, A.
\newblock Language models do not follow occam's razor: A benchmark for inductive and abductive reasoning, 2025.

\bibitem[Sun et~al.(2024)Sun, Xiao, Li, Ji, Chen, and Zhang]{sun-etal-2024-exploring}
Sun, Z., Xiao, Y., Li, J., Ji, Y., Chen, W., and Zhang, M.
\newblock Exploring and mitigating shortcut learning for generative large language models.
\newblock In Calzolari, N., Kan, M.-Y., Hoste, V., Lenci, A., Sakti, S., and Xue, N. (eds.), \emph{Proceedings of the 2024 Joint International Conference on Computational Linguistics, Language Resources and Evaluation (LREC-COLING 2024)}, pp.\  6883--6893, Torino, Italia, May 2024. ELRA and ICCL.

\bibitem[Wang et~al.(2025)Wang, Gao, Gu, Pu, Cui, Wei, Liu, Jing, Ye, Shao, et~al.]{wang2025internvl3_5}
Wang, W., Gao, Z., Gu, L., Pu, H., Cui, L., Wei, X., Liu, Z., Jing, L., Ye, S., Shao, J., et~al.
\newblock Internvl3.5: Advancing open-source multimodal models in versatility, reasoning, and efficiency.
\newblock \emph{arXiv preprint arXiv:2508.18265}, 2025.

\bibitem[Wang et~al.(2022)Wang, Wei, Schuurmans, Le, Chi, Narang, Chowdhery, and Zhou]{wang2022self}
Wang, X., Wei, J., Schuurmans, D., Le, Q., Chi, E., Narang, S., Chowdhery, A., and Zhou, D.
\newblock Self-consistency improves chain of thought reasoning in language models.
\newblock \emph{arXiv preprint arXiv:2203.11171}, 2022.

\bibitem[Wei et~al.(2023)Wei, Yuan, Yang, Shen, Li, Wang, and Chen]{wei-etal-2023-tackling}
Wei, Y., Yuan, S., Yang, R., Shen, L., Li, Z., Wang, L., and Chen, M.
\newblock Tackling modality heterogeneity with multi-view calibration network for multimodal sentiment detection.
\newblock In Rogers, A., Boyd-Graber, J., and Okazaki, N. (eds.), \emph{Proceedings of the 61st Annual Meeting of the Association for Computational Linguistics (Volume 1: Long Papers)}, pp.\  5240--5252, Toronto, Canada, July 2023. Association for Computational Linguistics.
\newblock \doi{10.18653/v1/2023.acl-long.287}.

\bibitem[Wu et~al.(2025{\natexlab{a}})Wu, Yang, Zhou, Fang, Song, Sun, and Ji]{wu2025grounded}
Wu, Q., Yang, X., Zhou, Y., Fang, C., Song, B., Sun, X., and Ji, R.
\newblock Grounded chain-of-thought for multimodal large language models.
\newblock \emph{arXiv preprint arXiv:2503.12799}, 2025{\natexlab{a}}.

\bibitem[Wu et~al.(2025{\natexlab{b}})Wu, Zhang, Yao, Du, Yan, Ding, Wu, and Li]{wu2025antidote}
Wu, Y., Zhang, L., Yao, H., Du, J., Yan, K., Ding, S., Wu, Y., and Li, X.
\newblock Antidote: A unified framework for mitigating lvlm hallucinations in counterfactual presupposition and object perception.
\newblock In \emph{Proceedings of the Computer Vision and Pattern Recognition Conference}, pp.\  14646--14656, 2025{\natexlab{b}}.

\bibitem[Wu et~al.(2025{\natexlab{c}})Wu, Huang, and Wu]{wu2025spurioussignalsdebiasingmultimodal}
Wu, Z., Huang, H.-Y., and Wu, Y.
\newblock Beyond spurious signals: Debiasing multimodal large language models via counterfactual inference and adaptive expert routing, 2025{\natexlab{c}}.

\bibitem[Ye et~al.(2023)Ye, Xu, Xu, Ye, Yan, Zhou, Wang, Hu, Shi, Shi, et~al.]{ye2023mplug}
Ye, Q., Xu, H., Xu, G., Ye, J., Yan, M., Zhou, Y., Wang, J., Hu, A., Shi, P., Shi, Y., et~al.
\newblock mplug-owl: Modularization empowers large language models with multimodality.
\newblock \emph{arXiv preprint arXiv:2304.14178}, 2023.

\bibitem[Yu et~al.(2020)Yu, Xu, Meng, Zhu, Ma, Wu, Zou, and Yang]{yu2020ch}
Yu, W., Xu, H., Meng, F., Zhu, Y., Ma, Y., Wu, J., Zou, J., and Yang, K.
\newblock Ch-sims: A chinese multimodal sentiment analysis dataset with fine-grained annotation of modality.
\newblock In \emph{Proceedings of the 58th annual meeting of the association for computational linguistics}, pp.\  3718--3727, 2020.

\bibitem[Zadeh et~al.(2016)Zadeh, Zellers, Pincus, and Morency]{zadeh2016mosimultimodalcorpussentiment}
Zadeh, A., Zellers, R., Pincus, E., and Morency, L.-P.
\newblock Mosi: Multimodal corpus of sentiment intensity and subjectivity analysis in online opinion videos, 2016.

\bibitem[Zadeh et~al.(2017)Zadeh, Chen, Poria, Cambria, and Morency]{zadeh2017tensor}
Zadeh, A., Chen, M., Poria, S., Cambria, E., and Morency, L.-P.
\newblock Tensor fusion network for multimodal sentiment analysis.
\newblock \emph{arXiv preprint arXiv:1707.07250}, 2017.

\bibitem[Zelikman et~al.(2024)Zelikman, Harik, Shao, Jayasiri, Haber, and Goodman]{zelikman2024quiet}
Zelikman, E., Harik, G., Shao, Y., Jayasiri, V., Haber, N., and Goodman, N.~D.
\newblock Quiet-star: Language models can teach themselves to think before speaking.
\newblock \emph{arXiv preprint arXiv:2403.09629}, 2024.

\bibitem[Zhang et~al.(2025)Zhang, Zeng, and Gu]{zhang2025simignore}
Zhang, X., Zeng, F., and Gu, C.
\newblock Simignore: Exploring and enhancing multimodal large model complex reasoning via similarity computation.
\newblock \emph{Neural Networks}, 184:\penalty0 107059, 2025.

\bibitem[Zhao et~al.(2025{\natexlab{a}})Zhao, Wei, and Bo]{zhao2025r1}
Zhao, J., Wei, X., and Bo, L.
\newblock R1-omni: Explainable omni-multimodal emotion recognition with reinforcement learning.
\newblock \emph{arXiv preprint arXiv:2503.05379}, 2025{\natexlab{a}}.

\bibitem[Zhao et~al.(2025{\natexlab{b}})Zhao, Yang, Peng, Bai, Yao, Sun, Chen, Fu, Wei, Bo, et~al.]{zhao2025humanomni}
Zhao, J., Yang, Q., Peng, Y., Bai, D., Yao, S., Sun, B., Chen, X., Fu, S., Wei, X., Bo, L., et~al.
\newblock Humanomni: A large vision-speech language model for human-centric video understanding.
\newblock \emph{arXiv preprint arXiv:2501.15111}, 2025{\natexlab{b}}.

\bibitem[Zhao et~al.(2025{\natexlab{c}})Zhao, Zheng, Li, and Liu]{zhao2025multimodal}
Zhao, K., Zheng, M., Li, Q., and Liu, J.
\newblock Multimodal sentiment analysis-a comprehensive survey from a fusion methods perspective.
\newblock \emph{IEEE Access}, 2025{\natexlab{c}}.

\bibitem[Zhao et~al.(2023)Zhao, Li, Joty, Qin, and Bing]{zhao2023verify}
Zhao, R., Li, X., Joty, S., Qin, C., and Bing, L.
\newblock Verify-and-edit: A knowledge-enhanced chain-of-thought framework.
\newblock \emph{arXiv preprint arXiv:2305.03268}, 2023.

\bibitem[Zhou et~al.(2023)Zhou, Cui, Yoon, Zhang, Deng, Finn, Bansal, and Yao]{zhou2023analyzing}
Zhou, Y., Cui, C., Yoon, J., Zhang, L., Deng, Z., Finn, C., Bansal, M., and Yao, H.
\newblock Analyzing and mitigating object hallucination in large vision-language models.
\newblock \emph{arXiv preprint arXiv:2310.00754}, 2023.

\bibitem[Zhu et~al.(2025)Zhu, Tao, Dong, and Xu]{zhu2025mitigating}
Zhu, Y., Tao, L., Dong, M., and Xu, C.
\newblock Mitigating object hallucinations in large vision-language models via attention calibration.
\newblock \emph{arXiv preprint arXiv:2502.01969}, 2025.

\end{thebibliography}

\newpage
\appendix
\onecolumn

\section{Implementation Details of Reward Mechanisms}
\label{app:impl}

\subsection{Reward Composition and Weights}
\label{app:reward-weights}
As introduced in Sec. 3.1, the total reward $R$ is calculated as the weighted sum of individual task-specific rewards. All reward components $r_k$ are normalized to $[0, 1]$ before scaling. Table~\ref{tab:reward-weights} lists the specific weights used during GRPO training. These weights were determined through a coarse grid search on a validation subset to balance logical rigour (citations/evidence) and task accuracy.

\begin{table}[h]
\centering
\caption{Hyperparameters for Hierarchical Reward Shaping.}
\label{tab:reward-weights}
\begin{tabular}{lll p{6cm}}
\toprule
\textbf{Category} & \textbf{Component} & \textbf{Weight} $\lambda_k$ & \textbf{Primary Optimization Goal} \\
\midrule
Performance & Accuracy ($r_{\text{acc}}$) & 4.0 & Maximize F1-score with length penalty $P_{\ell}$. \\
\midrule
Protocol & Consistency ($r_{\text{fmt}}$) & 0.5 & Ensure JSON validly and tag ordering. \\
         & Reasoning ($r_{\text{think}}$) & 0.5 & Enforce comparative/differential thinking blocks. \\
\midrule
Reliability & Citation ($r_{\text{cite}}$) & 1.0 & Encourage explicit hypothesis referencing. \\
            & Evidence ($r_{\text{evid}}$) & 2.0 & Enforce intra-trace consistency via fuzzy matching. \\
            & Grounding ($r_{\text{sem}}$) & 2.0 & Align reasoning with multimodal cue annotations. \\
\bottomrule
\end{tabular}
\end{table}

\subsection{Detailed Formulations of Reward Components}
\label{app:reward_detailed_formulations}
\textbf{Accuracy with Length Penalty.} 
To prevent reward hacking via verbosity, $r_{\text{acc}}$ incorporates a penalty $P_{\ell}$:
\begin{equation}
P_{\ell} = \frac{L}{L_{\text{pre}}}
\end{equation}
where $L$ is the number of emotion words in the ground truth labels and $L_{\text{pre}}$ is the number of emotion words in the model-predicted labels.

\textbf{Fuzzy Matching for $r_{\text{evid}}$.}
The $\mathrm{match}(c, V)$ operator (Eq. 4) handles surface variations by: (i) converting to lowercase, (ii) stripping punctuation, and (iii) verifying if the citation string $c$ exists as a substring within the evidence pool $V$. For ambiguous cases, a Levenshtein similarity threshold of $0.85$ is applied.

\textbf{Discretization of Semantic Grounding ($r_{\text{sem}}$).}
The function $Q(\cdot)$ in Eq. 5 maps continuous cosine similarity $s$ to discrete rewards to stabilize the RL gradient:
\begin{equation}
Q(s) = \begin{cases} 1.0, & s \geq 0.7 \\ 0.5, & 0.5 \leq s < 0.7 \\ 0, & s < 0.5 \end{cases}
\end{equation}
We use \texttt{all-MiniLM-L6-v2} as the backbone for $\text{emb}(\cdot)$.

\section{Data Preparation}
\label{app:data_preparation}

\begin{figure*}[ht]
  \vskip 0.2in
    \centerline{\includegraphics[width=0.9\textwidth]{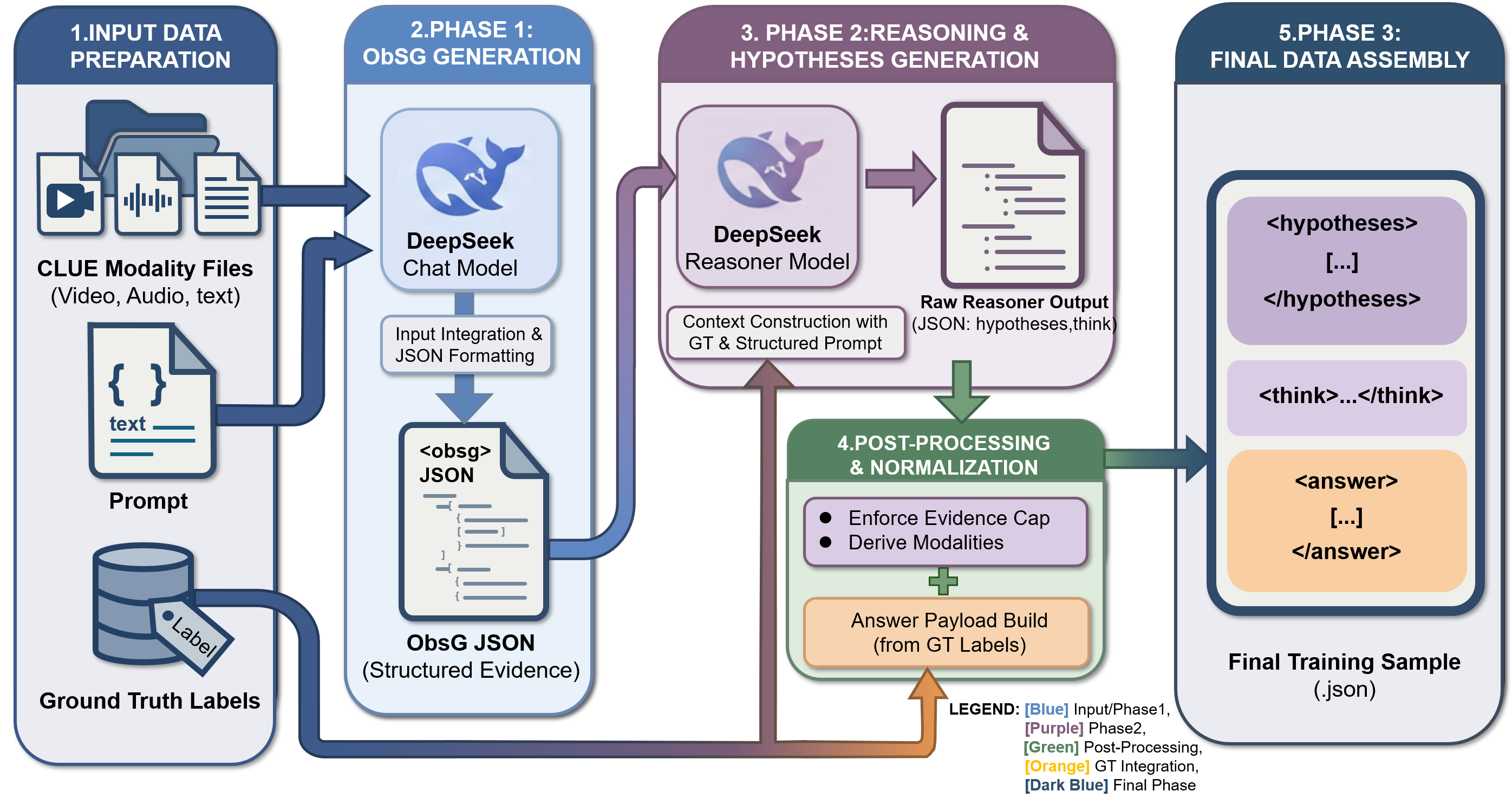}}
    \caption{
      Overview of the Generation stages of ObsG.
    }
    \label{fig:ObsG_generation}
\end{figure*}

This section details the technical implementation of the automated generation pipeline used to construct data for cold-start SFT and GRPO RL. The overall procedure, illustrated in Figure~\ref{fig:ObsG_generation} and formalized in Algorithm~\ref{alg:data_generation}, is designed as a structured, sequential workflow that transforms raw multimodal descriptions and Ground Truth (GT) labels into standardized training samples following the HyDRA output interface, i.e., a structured response composed of \texttt{<hypotheses>}, \texttt{<think>}, and \texttt{<answer>}.

The pipeline begins with input preparation, followed by Phase~1: ObsG generation. In Phase~1, a general-purpose DeepSeek Chat model converts unstructured multimodal text into structured ObsG JSON objects that summarize objective, time-sliced evidence. The core of the pipeline is Phase~2, where a specialized DeepSeek Reasoner model instantiates HyDRA’s Propose–Verify–Decide protocol. Specifically, given a composite context consisting of the objective ObsG and the GT labels, the reasoner is prompted to (i) generate multiple competing hypotheses in \texttt{<hypotheses>}, each grounded in modality-specific evidence candidates; (ii) perform evidence-constrained comparison in \texttt{<think>} by explicitly contrasting shared cues and conflicting cues across hypotheses and selecting the most coherent explanation; and (iii) output the final open-vocabulary emotion set in \texttt{<answer>}, consistent with the verified decision.

In Phase~3, we apply deterministic post-processing and normalization to ensure interface stability and reduce spurious variance. Finally, we assemble the verified HyDRA reasoning modules, the original observation modules, and the formatted target answers into a single structurally standardized JSON training sample, completing the automated construction of high-quality cold-start data into a single standardized JSON training sample, as shown in the output of Step 5 in Figure~\ref{fig:ObsG_generation}.

For completeness, we also reuse this pipeline to build GRPO training data. In this case, we only retain the Phase~1 outputs (ObsG JSON), and omit Phases~2–3, because GRPO requires objective observations as reward signals rather than fully instantiated HyDRA responses.

\begin{tcolorbox}[title=System Prompt: Cold-Start Data Generation]
You are a structured reasoning assistant. Use ONLY the provided ObservationGraph (ObsG).
\begin{itemize}
    \item \textbf{Input Context}: Multimodal evidence (video/audio/text) and EVAL-ONLY affect profiles (Ground Truth hints).
    \item \textbf{Constraint}: Do NOT quote EVAL-ONLY content directly. Rank evidence by cross-modality coverage.
    \item \textbf{Output Schema (JSON)}:
    \begin{itemize}
        \item[1.] $\texttt{hypotheses}$: A list of 1--3 hypotheses. Each must contain an $\texttt{assumption}$ and $\texttt{evidence\_ids\_top5}$ (unique IDs).
        \item[2.] $\texttt{think}$: A reasoning trace using **[Common]**, **[Differences]**, and **[Decision]** headers. **MANDATORY: Cite specific evidence like [v1][a2] and cross-reference hypotheses [H1]. [v1][a2] needed to be natural language descriptions in the raw data.**
    \end{itemize}
    \item \textbf{Termination}: Conclude the [Decision] section with "leading to the conclusion for an emotional state of [DESCRIPTOR]."
\end{itemize}
\end{tcolorbox}

\begin{algorithm}[tb]
  \caption{Cold-Start Data Generation Pipeline}
  \label{alg:data_generation}
  
  \renewcommand{\algorithmiccomment}[1]{// \textit{#1}} 
  
\begin{algorithmic}
  \STATE {\bfseries Input:} Multimodal Data $(V, A, T)$, Ground Truth $\text{GT}_{\text{labels}}$, Prompt Templates
  \STATE {\bfseries Output:} Structured Training Sample $S_{\text{final}}$

  \STATE \COMMENT{Phase 1: Observation Generation}
  \STATE $\text{Prompt}_{\text{ObsG}} \leftarrow \text{ConstructObsGPrompt}(V, A, T, \text{Template})$
  \STATE $\text{ObsG}_{\text{JSON}} \leftarrow \text{LLM}_{\text{Chat}}(\text{Prompt}_{\text{ObsG}})$

  \STATE \vspace{0.2em}
  \STATE \COMMENT{Phase 2: Reasoning \& Hypotheses Generation}
  \STATE $\text{Context} \leftarrow \text{EmbedGT}(\text{ObsG}_{\text{JSON}}, \text{GT}_{\text{labels}})$
  \STATE $\text{Prompt}_{\text{Reason}} \leftarrow \text{ConstructReasonerPrompt}(\text{Context})$
  \STATE $\text{Raw}_{\text{Output}} \leftarrow \text{LLM}_{\text{Reasoner}}(\text{Prompt}_{\text{Reason}})$
  \STATE $\text{Hypotheses, CoT} \leftarrow \text{Parse}(\text{Raw}_{\text{Output}})$

  \STATE \vspace{0.2em}
  \STATE \COMMENT{Phase 3: Post-Processing \& Final Assembly}
  \STATE $\text{Hypotheses}' \leftarrow \text{EnforceEvidenceCap}(\text{Hypotheses}, \text{top\_k}=5)$
  \STATE $\text{Hypotheses}'' \leftarrow \text{DeriveModalities}(\text{Hypotheses}')$
  \STATE $\text{Payload}_{\text{Answer}} \leftarrow \text{FormatGT}(\text{GT}_{\text{labels}})$
  \STATE $S_{\text{final}} \leftarrow \text{AssembleXML}(\dots)$ 

  \STATE {\bfseries return} $S_{\text{final}}$
\end{algorithmic}
\end{algorithm}

\section{Datasets Details}
\label{app:datasets}

We evaluate our method on a diverse set of public multimodal affect benchmarks (Table~\ref{tab:dataset_card}). For Sentiment Analysis, we use the official test splits of CMU-MOSI and CH-SIMS. For Basic Emotion Recognition, we use the test sets of MER2023 (MER-MULTI) and MER2024 (MER-SEMI). For Fine-grained Emotion Detection, we use the MER-FG test set to assess open-vocabulary fine-grained emotion prediction. Dataset splits and sample counts are summarized in Table~\ref{tab:dataset_card}.

\begin{table}[t]
  \caption{Dataset card of the benchmarks used in our experiments. We group datasets by task type and report the official split adopted for evaluation.}
  \label{tab:dataset_card}
  \begin{center}
    \begin{small}
      \begin{sc}
        \begin{tabularx}{\columnwidth}{l X c c}
          \toprule
          Dataset Type & Raw Dataset & Selected Subset & \# Samples \\
          \midrule
          \multirow{2}{*}{Sentiment Analysis}
            & CMU-MOSI~\cite{zadeh2017tensor} & Test & 686 \\
            & CH-SIMS~\cite{yu2020ch}         & Test & 457 \\
          \midrule
          \multirow{2}{*}{Basic Emotion Recognition}
            & MER2023~\cite{10.1145/3581783.3612836} & MER-MULTI (Test) & 411 \\
            & MER2024~\cite{lian2024mer}             & MER-SEMI (Test)  & 1,169 \\
          \midrule
          Fine-grained Emotion Detection
            & MER-FG~\cite{lian2025mer} & Test & 1,200 \\
          \bottomrule
        \end{tabularx}
      \end{sc}
    \end{small}
  \end{center}
  \vskip -0.1in
\end{table}

\section{Baseline Details}
\label{app:baselines}

This section summarizes the backbone components of the MLLMs referenced in our experiments and comparisons. 

\paragraph{HumanOmni.} We adopt HumanOmni-0.5B \cite{zhao2025humanomni} as our backbone, which is a large multimodal model tailored for human-centric tasks via a dual-path vision architecture. HumanOmni is offered in two scales: 0.5B and 7B. While both versions share the same fundamental design, they differ in component scales: HumanOmni-0.5B integrates a SigLip-400M vision tower, a HumanVision-Base encoder, and a Qwen2-0.5B LLM, whereas the 7B version scales these to SigLip-SO400M, HumanVision-Large, and Qwen2-7B, respectively. To maximize the demonstration of our method's effectiveness under constrained computational resources, we employ the 0.5B variant, which provides high-fidelity human-centric features within a compact parameter space.

\paragraph{Otter.}
Otter is a Flamingo-style, instruction-tuned multimodal assistant that supports in-context multimodal instruction following, and we include it as a generic image-based MLLM baseline. \cite{li2025otter}

\paragraph{Video-LLaVA.}
Video-LLaVA extends LLaVA-style instruction tuning to jointly handle images and videos via a unified visual representation, serving as a strong video-chat baseline without explicit audio modeling. \cite{lin2024video}

\paragraph{Video-ChatGPT.}
Video-ChatGPT is a video conversation model that pairs a video-adapted visual encoder with an LLM for detailed video dialogue, and we use it as a representative early video-chat baseline. \cite{maaz2024video}

\paragraph{LLaMA-Vid.}
LLaMA-Vid targets long-context video understanding by reducing per-frame visual token length, and we adopt it to test whether improved temporal scalability benefits affect reasoning. \cite{li2024llama}

\paragraph{Chat-UniVi.}
Chat-UniVi unifies image and video understanding with dynamic visual tokens under a unified visual representation, and we treat it as a competitive unified image-video MLLM baseline. \cite{jin2024chat}

\paragraph{PandaGPT.} 
PandaGPT utilizes the multimodal joint representation space of ImageBind to align diverse inputs with the Vicuna language model, serving as a versatile baseline for general-purpose multimodal instruction following\cite{su2023pandagptmodelinstructionfollow}.

\paragraph{VideoChat.}
VideoChat is a chat-centric video understanding framework that connects video foundation models with LLMs via a learnable interface, and we include it as a widely used video understanding baseline. \cite{li2025videochat}

\paragraph{mPLUG-Owl.}
mPLUG-Owl equips LLMs with multimodality via modularized training and multimodal instruction tuning, and we use it as a representative modular MLLM baseline. \cite{ye2023mplug}

\paragraph{R1-Omni.}
R1-Omni applies reinforcement learning with verifiable rewards to improve omni-multimodal (audio-visual) emotion recognition and explainability, making it a key affect-specialized baseline. It is based on HumanOmni-0.5B. \cite{zhao2025r1}

\paragraph{AffectGPT (OV-MERD).}
We use AffectGPT(OV-MERD) as specified in the MER2025 setting, i.e., the AffectGPT variant trained on OV-MERD for open-vocabulary/fine-grained emotion understanding comparisons. \cite{lian2025mer,lian2025affectgpt}

\section{Emotion Wheel-based (EW) Metric}
\label{app:ew_metric}

\begin{figure}[ht]
    \centering
    \begin{subfigure}[b]{0.3\textwidth}
        \centering
        \includegraphics[width=\textwidth]{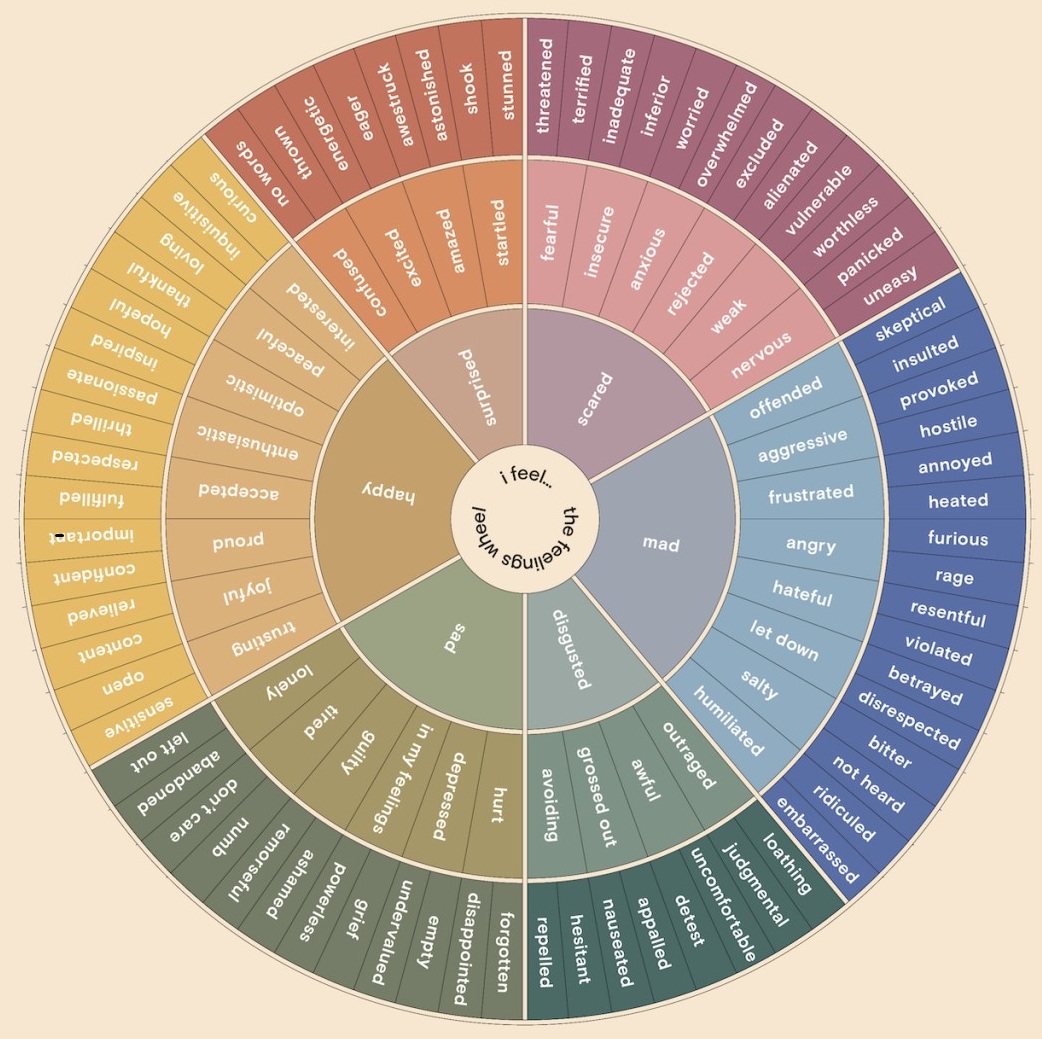}
        \caption{w1}
        \label{fig:wheel1}
    \end{subfigure}
    \hfill 
    \begin{subfigure}[b]{0.3\textwidth}
        \centering
        \includegraphics[width=\textwidth]{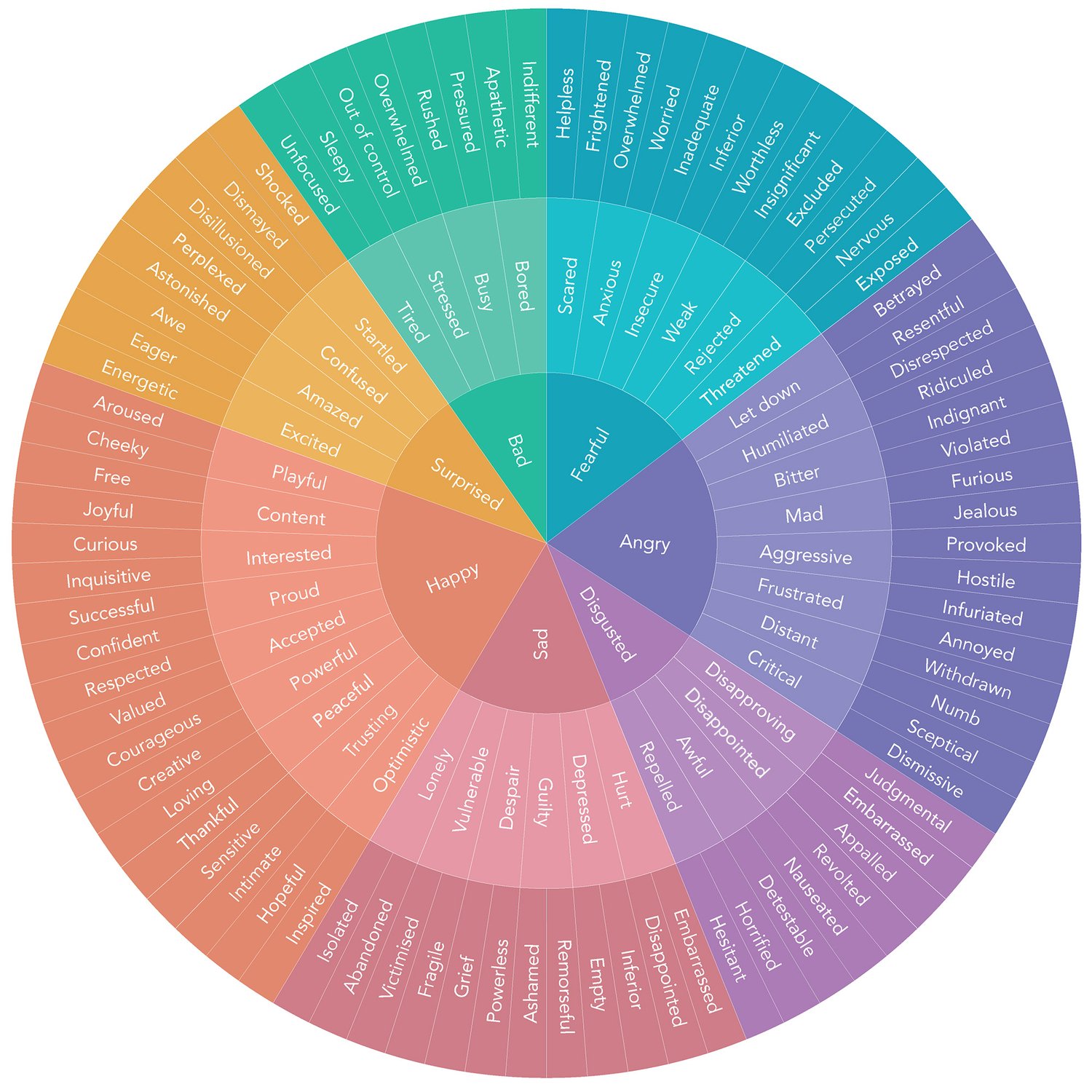}
        \caption{w2}
        \label{fig:wheel2}
    \end{subfigure}
    \hfill
    \begin{subfigure}[b]{0.3\textwidth}
        \centering
        \includegraphics[width=\textwidth]{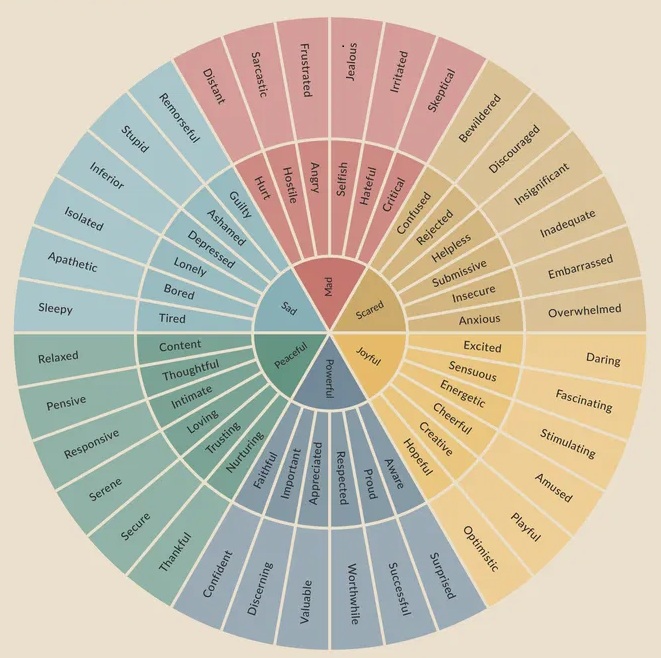}
        \caption{w3}
        \label{fig:wheel3}
    \end{subfigure}

    \vspace{1em} 

    \begin{subfigure}[b]{0.3\textwidth}
        \centering
        \includegraphics[width=\textwidth]{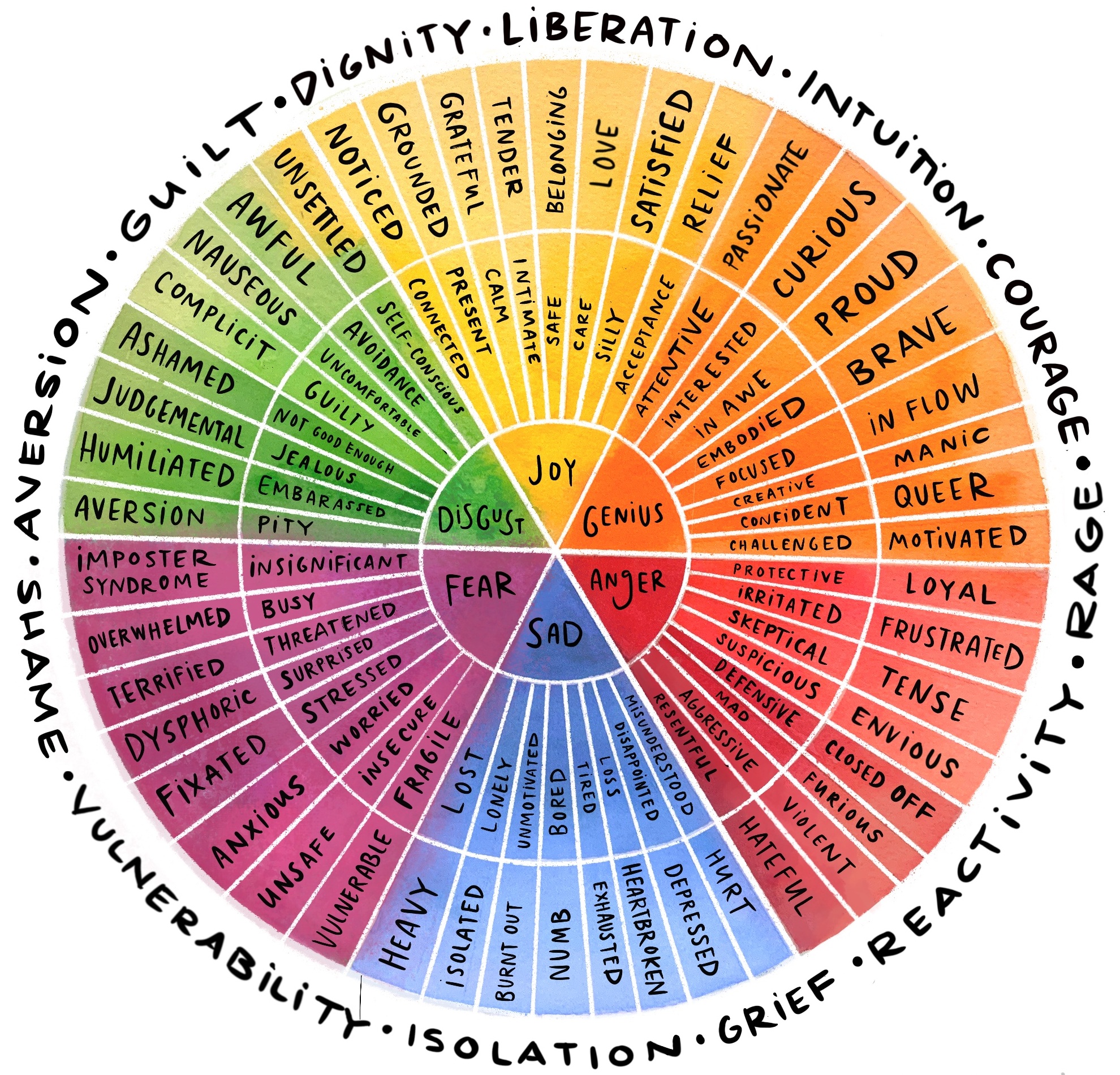}
        \caption{w4}
        \label{fig:wheel4}
    \end{subfigure}
    \hspace{2em} 
    \begin{subfigure}[b]{0.3\textwidth}
        \centering
        \includegraphics[width=\textwidth]{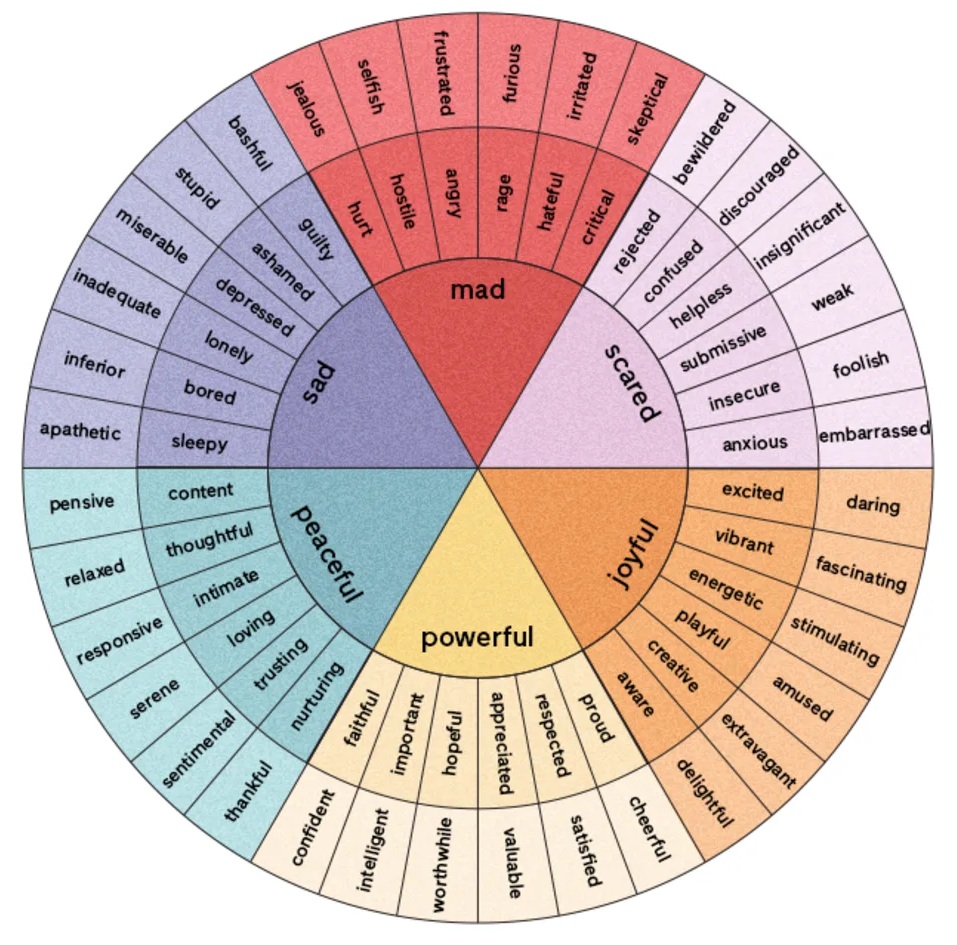}
        \caption{w5}
        \label{fig:wheel5}
    \end{subfigure}

    \caption{Emotion wheel. We use five emotion wheels, all of which are derived from previous research~\cite{lian2025affectgpt}.}
    \label{fig:emotion_wheels}
\end{figure}

To provide a robust evaluation for open-vocabulary emotion prediction, we adopt the \textit{Emotion Wheel-based (EW) metric} proposed in prior literature~\cite{lian2025affectgpt}. The illustrations are Figure~\ref{fig:emotion_wheels}. This metric accounts for semantic redundancies and synonyms by employing a hierarchical clustering strategy and set-level evaluation. The computation process is objective and primarily consists of two components: neutralizing synonym influence and defining set-level metrics.

\subsection{Handling Synonyms and Variations}
To mitigate the impact of morphological variations and semantic overlaps, a three-level hierarchical grouping strategy is employed, represented by a composite clustering function $G_{w_k}(\cdot)$:

\begin{itemize}
    \item \textbf{(L1) Morphological Normalization ($F_{l_1}$):} This function maps different forms of words to their base form. For instance, words like \textit{happier} and \textit{happiness} are normalized to \textit{happy}.
    \item \textbf{(L2) Synonym Mapping ($F_{l_2}$):} This mapping function unifies synonyms into a single representative word. For example, \textit{joyful} and \textit{cheerful} are mapped to \textit{happy}.
    \item \textbf{(L3) Emotion Wheel Clustering ($F_{l_3}^{w_k}$):} This function maps outer, more specific labels to their corresponding inner fundamental labels. Following \cite{lian2025affectgpt}, $K=5$ different emotion wheels are adopted to ensure comprehensive coverage.
\end{itemize}

The complete clustering operation for a specific wheel $w_k$ is summarized as:
\begin{equation}
    G_{w_k}(\cdot) = F_{l_3}^{w_k}\left(F_{l_2}\left(F_{l_1}(\cdot)\right)\right), \quad k \in [1, K].
\end{equation}

\subsection{Set-level Metric Calculation}
Given that the number of predicted and annotated emotions may vary across samples, the evaluation is conducted using set-level metrics. Suppose the dataset contains $N$ samples. For a sample $x_i$, let $\mathbf{Y}_i = \{y_i^j\}_{j=1}^{n_i}$ denote the set of true labels and $\hat{\mathbf{Y}}_i = \{\hat{y}_i^j\}_{j=1}^{\hat{n_i}}$ denote the set of predicted labels. The metrics for each wheel $k$ are defined as:

\begin{equation}
    \text{Precision}_S^k = \frac{1}{N} \sum_{i=1}^N \frac{|G_{w_k}(\mathbf{Y}_i) \cap G_{w_k}(\hat{\mathbf{Y}}_i)|}{|G_{w_k}(\hat{\mathbf{Y}}_i)|},
\end{equation}

\begin{equation}
    \text{Recall}_S^k = \frac{1}{N} \sum_{i=1}^N \frac{|G_{w_k}(\mathbf{Y}_i) \cap G_{w_k}(\hat{\mathbf{Y}}_i)|}{|G_{w_k}(\mathbf{Y}_i)|},
\end{equation}

\begin{equation}
    F_S^k = 2 \times \frac{\text{Precision}_S^k \times \text{Recall}_S^k}{\text{Precision}_S^k + \text{Recall}_S^k}.
\end{equation}

In these set-wise operations, duplicate emotion words are automatically removed.

\subsection{Final Score Computation}
Finally, the comprehensive EW score is determined by calculating the average F1-score across all $K$ emotion wheels:
\begin{equation}
    \text{EW}(\mathbf{Y}_i, \hat{\mathbf{Y}}_i) = \frac{1}{K} \sum_{k=1}^K F_S^k.
\end{equation}

\label{app:ew}

\begin{figure}[ht]
  \vskip 0.2in
  \begin{center}
    \centerline{\includegraphics[width=0.7\textwidth]{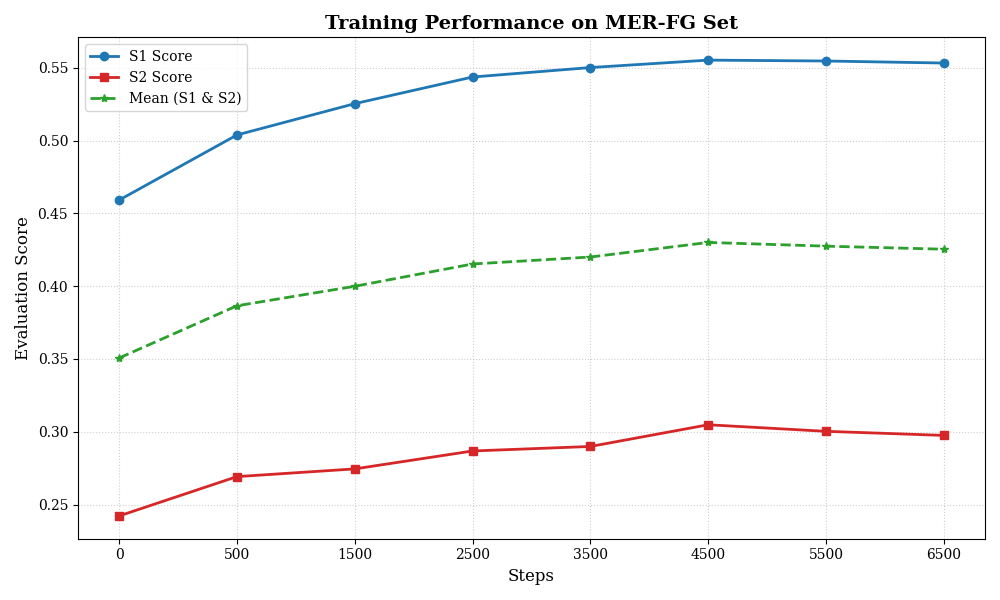}}
    \caption{      
Evaluation scores across various epochs on the MER-FG.
}    
\label{fig:different-epoch}  
\end{center}
\end{figure}

\section{Experimental Setup}
\label{app:experimental_setup}

\textbf{Training Details.}
Our models were trained on a cluster of up to 8 NVIDIA L20 GPUs. To balance computational efficiency with memory constraints, we uniformly sampled 8 frames for video inputs during training. All experiments were conducted using BF16 precision and FlashAttention-2, with DeepSpeed ZeRO-3 employed for memory optimization.The training process consists of two stages: SFT followed by GRPO RL. 

\begin{itemize}
    \item \textbf{SFT Stage:} We use BF16 training with gradient checkpointing and FlashAttention. The maximum sequence length is 2048, the per-device batch size is 2 with gradient accumulation to a global batch of 16. We train for 5 epochs using a cosine schedule, learning rate $2\times 10^{-4}$, warmup ratio 0.01, and weight decay 0.0. The multimodal configuration uses a vision tower and an audio tower, a lightweight projector, and enables modality start/end wrappers; the number of video frames is 8.
    \item \textbf{GRPO RL Stage:} We run BF16 training with gradient checkpointing and FlashAttention. The prompt and completion lengths are 1024/1024. We sample $G=8$ completions per prompt, use a per-device batch size of 1 with gradient accumulation of 4, and train for 4.5k steps. This duration is empirically selected as the optimal convergence point; as shown in Figure~\ref{fig:different-epoch}, the evaluation scores (e.g., $S_1$ and Mean) peak at approximately 4.5k steps and plateau thereafter. The learning rate is $1\times 10^{-6}$ with warmup ratio 0.03. The maximum pixel budget for visual inputs is capped at 401408.
\end{itemize}

\section{Limitations}
\label{app:limitations}

\textbf{Model scale and compute.}
We train and evaluate HyDRA on a relatively small backbone (0.5B), and do not perform RL training on 7B-scale (or larger) MLLMs. This choice is primarily constrained by available compute, and it may limit the absolute performance ceiling as well as the generality of our conclusions across model sizes. Larger backbones could better absorb the Propose--Verify--Decide objective, support stronger perception and longer-context reasoning, and potentially change the optimal trade-offs (e.g., between diversity and verification strictness). With access to larger-scale training resources, we plan to scale HyDRA to 7B+ models and systematically study scaling behavior under the same evaluation protocol.

\textbf{Sensitivity to cold-start formatting and fixed hypothesis budget.}
HyDRA is strongly shaped by the cold-start training format and the structured rationale schema (e.g., the explicit \texttt{<hypotheses>} field). In our current setting, generating two hypotheses is the best-performing configuration; however, this fixed budget may not be robust when the input has higher information density, more complex situational dynamics, or richer multi-label affective mixtures. In such cases, two candidates may fail to cover the space of plausible explanations, weakening the comparative verification stage and re-introducing premature commitment. Future work will explore \emph{elastic} hypothesis generation—adapting the number (and granularity) of hypotheses to the uncertainty and cue diversity of each instance—while keeping verification efficient (e.g., early stopping, hypothesis pruning, or learned controllers).

\textbf{Backbone perception as an upstream bottleneck.}
Our contribution targets the \emph{reasoning paradigm} (how the model forms and adjudicates interpretations), but OV-MER performance can be fundamentally limited by the base MLLM's perceptual competence. If the model fails to reliably encode fine-grained visual or acoustic cues (i.e., it cannot ``see/hear'' them), then richer reasoning may not recover the missing evidence, and verification may become brittle or under-informed. This is especially relevant for subtle facial micro-expressions, prosody, and cross-modal timing. An important next step is to re-evaluate HyDRA on stronger multimodal backbones and perception modules to disentangle gains from improved perception versus improved abductive adjudication.

\textbf{Compatibility with emerging MLLM design directions.}
Recent MLLM research suggests two complementary directions for improving multimodal understanding: (i) decoupling the Perception--Cognition loop into explicit perception-first pipelines (often yielding large gains in recognition), and (ii) strengthening multimodal fusion directly in latent space to improve grounding and reduce reliance on language priors. Our current implementation does not explicitly incorporate either paradigm, and therefore may not realize the best achievable synergy between perception, fusion, and abductive reasoning. In future work, we will investigate combining HyDRA with perception-first/cognition-second workflows and with latent fusion architectures, to test whether stronger base representations further amplify the benefits of Propose--Verify--Decide training and evidence-closed verification.

\section{Analysis of Emotional Label Cardinality Distribution}
\label{app:length_penalty}
\begin{figure}[ht]
  \vskip 0.2in
  \begin{center}
    \centerline{\includegraphics[width=0.7\textwidth]{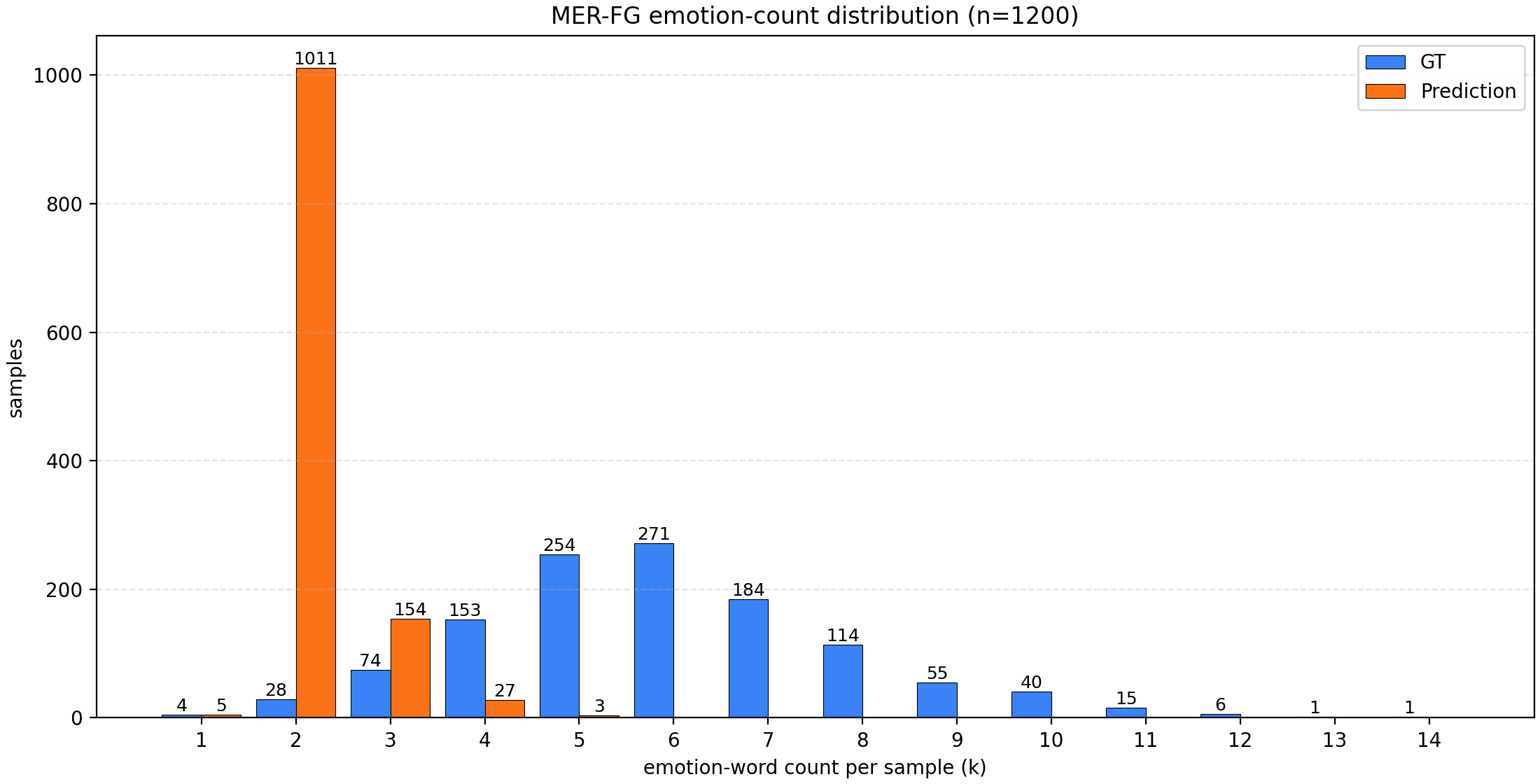}}
    \caption{      
Analysis of Emotional Label Cardinality Distribution
}    
\label{fig:emotion-count-distribution}  
\end{center}
\end{figure}

Figure~\ref{fig:emotion-count-distribution} illustrates the distribution of emotion-word counts per sample for both ground truth (GT) and model predictions. A common risk in open-vocabulary reward design is "label dumping," where the model hacks the reward by over-predicting labels to increase recall. However, our distribution shows that while GT labels typically range between 5 and 8, model predictions remain highly concentrated at $k=2$. This significant difference demonstrates that the length-penalty $P_{\ell}$ effectively constrains the model, ensuring that performance gains stem from semantic precision rather than cardinality-based reward hacking. Notably, this conciseness does not compromise accuracy; as detailed in Appendix~\ref{app:ew_metric}, our hierarchical metric maps predictions into unified taxonomic levels, allowing a few high-precision labels to effectively capture the primary emotional dimensions within the GT.

\end{document}